\documentclass[sigconf,prologue,table]{acmart}

\usepackage{amsmath}
\usepackage{amsfonts}  

\usepackage{graphicx}
\usepackage{textcomp}
\usepackage{xspace}
\usepackage{algorithm}
\usepackage{algpseudocode}
\usepackage[utf8]{inputenc} 
\usepackage[T1]{fontenc}    
\usepackage{hyperref}
\usepackage{booktabs}       
\usepackage{nicefrac}       
\usepackage{microtype}      
\usepackage{microtype}      
\usepackage{tabularx}
\usepackage{wrapfig}
\usepackage{multirow}
\usepackage{epsfig}
\usepackage{subfigure}
\usepackage{marvosym}
\usepackage{color}
\usepackage{threeparttable}
\usepackage{algorithm}
\usepackage{algpseudocode}
\usepackage{setspace}
\usepackage{amsthm}
\usepackage{pifont}
\newcommand{\cmark}{\ding{51}}
\newcommand{\xmark}{\ding{55}}

\usepackage[normalem]{ulem}

\AtBeginDocument{%
  }

\begin{document}

\title{FinKario: Event-Enhanced Automated Construction of Financial Knowledge Graph}


\author{Xiang Li}
\email{xli906@connect.hkust-gz.edu.cn}
\authornote{Equal Contribution.}
\affiliation{%
  \institution{The Hong Kong University of Science and Technology (Guangzhou)}
  \city{Guangzhou}
  \country{China}
  \postcode{511458}
}

\author{Penglei Sun}
\email{psun012@connect@hkust-gz.edu.cn}
\authornotemark[1]
\affiliation{%
  \institution{The Hong Kong University of Science and Technology (Guangzhou)}
  \city{Guangzhou}
  \country{China}
  \postcode{511458}
}

\author{Wanyun Zhou}
\email{wzhou266@connect.hkust-gz.edu.cn}
\affiliation{%
  \institution{The Hong Kong University of Science and Technology (Guangzhou)}
  \city{Guangzhou}
  \country{China}
  \postcode{511458}
}

\author{Zikai Wei}
\email{weizikai@idea.edu.cn}
\affiliation{%
  \institution{International Digital Economy Academy}
  \city{Shenzhen}
  \country{China}
  \postcode{518000}
}

\author{Yongqi Zhang}
\email{yongqizhang@hkust-gz.edu.cn}
\affiliation{%
  \institution{The Hong Kong University of Science and Technology (Guangzhou)}
  \city{Guangzhou}
  \country{China}
  \postcode{511458}
}
\authornote{Corresponding author.}

\author{Xiaowen Chu}
\email{xwchu@hkust-gz.edu.cn}
\affiliation{%
  \institution{The Hong Kong University of Science and Technology (Guangzhou)}
  \city{Guangzhou}
  \country{China}
  \postcode{511458}
}
\authornotemark[2]

\renewcommand{\shortauthors}{Xiang Li et al.}

\begin{abstract}

Individual investors are significantly outnumbered and disadvantaged in financial markets, overwhelmed by abundant information and lacking professional analysis.
Equity research reports stand out as crucial resources, offering valuable insights. By leveraging these reports, large language models (LLMs) can enhance investors' decision-making capabilities and strengthen financial analysis.
However, two key challenges limit their effectiveness: (1) the rapid evolution of market events often outpaces the slow update cycles of existing knowledge bases, (2) the long-form and unstructured nature of financial reports further hinders timely and context-aware integration by LLMs.
To address these challenges, we tackle both data and methodological aspects.
First, we introduce the Event-Enhanced Automated Construction of Financial Knowledge Graph \textbf{(FinKario)}, a dataset comprising over $305,360$ entities, $9,625$ relational triples, and $19$ distinct relation types. FinKario automatically integrates real-time company fundamentals and market events through prompt-driven extraction guided by professional institutional templates, providing structured and accessible financial insights for LLMs.
Additionally, we propose a Two-Stage, Graph-Based retrieval strategy \textbf{(FinKario-RAG)}, optimizing the retrieval of evolving, large-scale financial knowledge to ensure efficient and precise data access. 
Extensive experiments show that FinKario with FinKario-RAG achieves superior stock trend prediction accuracy, outperforming financial LLMs by \textbf{18.81\%} and institutional strategies by \textbf{17.85\%} on average in backtesting. 
\end{abstract}




\maketitle

\section{Introduction}

\begin{figure}[t]
    \centering
    \includegraphics[width=1.0\linewidth,
    trim=2pt 92pt 2pt 82pt, 
    clip
    ]{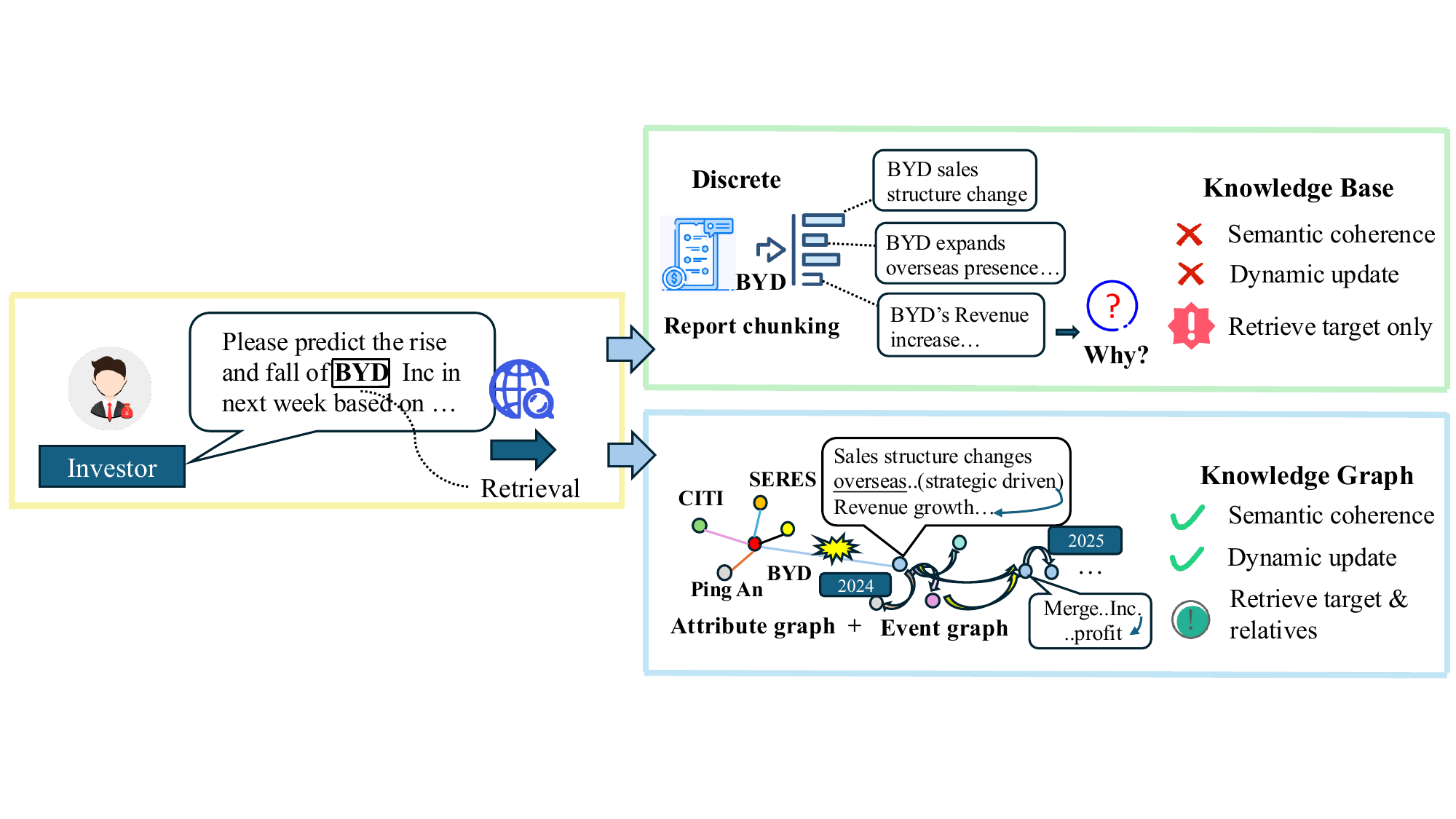}
    \caption{Comparison of Traditional Single-research Report Retrieval vs. FinKario Retrieval.}
    \label{fig:data retrieval comparison}
\end{figure}

The financial market is dominated by institutional investors, leaving individual investors at a significant disadvantage \cite{davis1996role,schlachter2013you}. 
Individual investors yet often struggle to make informed decisions due to a lack of access to professional-grade analysis~\cite{hilton2001psychology}. 
Equity research reports, which provide expert insights into market trends and company performance, serve as a critical resource to bridge this gap~\cite{zhou2024finrobot,kapellas2017financial}. 
LLMs can reason through complex financial data and interpret context~\cite{yang2023fingptopensourcefinanciallarge,wu2023bloomberggptlargelanguagemodel,araci2019finbert}. 
These capabilities make them suited for the analysis of financial research reports, thereby enabling timely and scalable insights.
Recent advances at the intersection of data mining and financial trading~\cite{wang2025pre,zhang2024multimodal} further highlight the potential of LLMs in this domain.



However, LLM-based research report analysis faces two primary challenges: 
\textbf{Ch1}, the rapid evolution of financial events \cite{cheng2020,xu2025finripple}, which outpaces the slow update cycles of existing knowledge bases shown in Table~\ref{tab:comparison}; and 
\textbf{Ch2}, the inherent complexity of processing long-form, unstructured data \cite{xia2024fetilda,sarmah2024hybridrag}.
Events such as earnings releases, product launches, and regulatory changes are key drivers of market behavior and are essential for understanding temporal shifts in asset performance \cite{mackinlay1997event,thompson1995empirical}.
Financial markets are characterized by a continuous stream of such events, which are often reflected in research reports. Yet, both traditional knowledge bases and modern LLMs struggle to keep pace with these dynamic updates.
As illustrated in Figure~\ref{fig:data retrieval comparison} and Table~\ref{tab:comparison}, two key limitations persist. 
\textbf{Lim1}: Current knowledge base approaches rely heavily on static, target-only retrieval and report chunking strategies, which lack semantic coherence, fail to provide contextual explanations (i.e., "why"), and cannot incorporate evolving market signals in real time.
\textbf{Lim2}: Existing event-centric Financial Knowledge Graphs (FKGs), for instance, still depend on manual or semi-automated construction pipelines \cite{zhu2023t,li2024findkg,kertkeidkachorn2023finkg,colombo2025template}, resulting in outdated representations that limit their effectiveness in dynamic environments.
Although LLMs excel at processing natural language, their internal 
knowledge remains static and infrequently updated, leading to persistent knowledge lag \cite{singh2024finqapt,adewale2023big}.


To address the mentioned challenge, we propose \textbf{FinKario}, a dual-structured financial knowledge graph built from equity research reports that can capture financial attributes, indicators and events automatically and dynamically (\textbf{Ch1}). 
It comprises two subgraphs: an \textbf{attribute subgraph} for stable fundamentals and an \textbf{event subgraph}, which captures time-sensitive events like quarterly financial performance and key profitability drivers (\textbf{Ch2}).
The process begins with the automated generation of schema for both subgraphs, using prompt-driven extraction guided by professional institutional frameworks such as the CFA handbook (\textbf{Lim2}). 
Specifically, the schema for the event subgraph adopts a top-down structure, guided by high-level categories extracted from academic reports provided by the University of Wisconsin, and is further refined into a detailed event ontology based on the Financial Industry Business Ontology (FIBO).
Based on these schemas, the pipeline further extracts structured knowledge from financial research reports using LLMs aligned with domain-specific templates.
Moreover, a quality control module ensures the reliability of the extracted knowledge by correcting erroneous or outdated information, normalizing entities, and completing missing attributes with the support of the Tushare financial data platform.
We collect a corpus of research reports from August 2024 to March 2025. 
The FinKario instance comprises over $305,360$ entities, $9,625$ relational triples, and $19$ distinct relation types.
In addition to the FinKario dataset, to address the retrieval challenge posed by dynamically evolving, large-scale financial knowledge, we propose the \textbf{FinKario-RAG}, which first retrieves information directly related to the queried entity, then expands retrieval to related entities and relationships, ensuring a holistic financial context essential for accurate predictions (\textbf{Lim1}).


To evaluate the effectiveness of FinKario, we conduct backtests comparing the predictive accuracy of our method with traditional financial LLMs and institutional strategies. 
Our results show that, on average, FinKario, combined with FinKario-RAG, outperforms existing financial LLMs by \textbf{$18.81 \%$} and institutional strategies by \textbf{$17.85 \%$} in predictive accuracy. 
Our ablation studies further verify that FinKario-RAG surpasses existing mainstream retrieval methods by an average of \textbf{$12.70\%$} in predictive accuracy.
The principal contributions of this paper are as follows:

\begin{itemize}
\item 
We introduce \textbf{FinKario}, a dynamic, event-driven financial knowledge graph with over \underline{305,360} entities, \underline{9,625} relational triples, and \underline{19} relation types, supporting automated updates without manual intervention or predefined domain knowledge, and enabling professional template–driven schema construction.

\item 
We propose \textbf{FinKario-RAG}, a retrieval strategy that integrates both industry-level and index-level perspectives to transcend the limitations of single-target retrieval, facilitating holistic and realistic analysis in line with practical investment scenarios and supporting retrieval over large-scale, dynamically evolving financial knowledge.

\item We empirically validate FinKario-RAG through extensive experiments, demonstrating that our method surpasses the runner-up by \textbf{58.14\%} in Sharpe ratio, \textbf{30.86\%} in Accumulative rate of return, and \textbf{1.04\%} in predictive accuracy, proving its effectiveness for financial analysis and stock trend forecasting.
\end{itemize}

\begin{table*}[t]
    \centering
    \caption{Comparison of key characteristics of financial knowledge graph construction methods, including events, dynamics, automation, major sources, and the number of entities, relations, and triples.}
    \label{tab:comparison}
    \renewcommand{\arraystretch}{1.02}
    \setlength{\tabcolsep}{4.5pt}
    \resizebox{.85\textwidth}{!}{
    \begin{tabular}{l c c c c c c c c}
        \toprule
        \textbf{Knowledge}    & \textbf{Event} & \textbf{Dynamic Updated} & \textbf{Automation} & \textbf{Entities} & \textbf{Relations} & \textbf{Triples} & \textbf{Major Source}\\
        \midrule
        FR2KG \cite{wang2021data}  & \xmark & \xmark & \xmark & 17,799    & 13   & 1,328 & Research Report \\
        KGEEF ~\cite{cheng2020} & \cmark & \xmark & \xmark & 5,262,423    & /  & 325,786 & News \\
        T-FinKB~\cite{zhu2023t}   & \xmark & \cmark & \xmark & 3,974     & 16   & / & News \\
        FinKG~\cite{kertkeidkachorn2023finkg}  & \xmark & \xmark & \xmark & 37,382,905 & 12  & 30M+ &  Market data      \\
        FEEKG~\cite{liu2024risk}     & \cmark & \xmark & \xmark & 112,000    & 12   & / & News    \\
        FinDKG~\cite{li2024findkg}   & \xmark & \cmark & \xmark & 13,645    & 15  & /  & News      \\
        FMAG~\cite{chen2024knowledge}  & \xmark & \cmark & \xmark & 8,052     & 5    & 5,664 & Research Report    \\
        FNRKPL~\cite{sun2025knowledge}  & \xmark & \xmark & \xmark & 11,432     & 6  & 52,384 & News    \\
        EKG~\cite{colombo2025template}  & \xmark & \xmark & \xmark & /     & 9  & / & Market data    \\
        FinRipple~\cite{xu2025finripple}  & \cmark & \xmark & \xmark & /     & 4  & / & News    \\
        \textbf{FinKario (Ours)}               & \cmark & \cmark & \cmark &         305,360  &       19  & 9,625 & Research Report  \\
        \bottomrule
    \end{tabular}}
    \vskip -0.15in
\end{table*}

\section{Related Work}

\subsection{Financial Knowledge Graph}

In recent years, Financial Knowledge Graphs (FKGs) have gained growing attention for their ability to improve financial data analysis and decision-making. Traditional approaches largely relied on standard natural language processing techniques, including semantic recognition, classification, and Named Entity Recognition (NER). For instance, Wang et al. \cite{wang2021data} proposed datasets and evaluations specifically for the construction of financial knowledge graphs, facilitating comprehensive assessments. Similarly, Liu et al. \cite{liu2024risk} introduced an approach focusing on financial event evolution through knowledge association, providing systematic risk identification and management.
In contrast, recent methodologies have begun leveraging large language models (LLMs) to summarize domain-specific corpora or participate directly in constructing relational entities within financial knowledge graphs.  Chen et al. \cite{chen2024knowledge} introduced a retrieval-augmented framework that incorporates structured financial knowledge graphs into language models to enhance reliability and accuracy in financial market analyses and report generation. Furthermore, Sun et al. \cite{sun2025knowledge} proposed a knowledge-enhanced prompt learning framework specifically aimed at improving financial news recommendation, leveraging FKGs to provide context-aware and accurate recommendations. 
However, most existing methods still depend on predefined schemas and manual input, underscoring the need for fully automated, schema-independent FKG construction in future research.

\subsection{Automatic Knowledge Graph Construction}

Recent advances have explored Automatic Knowledge Graph Construction (AKGC) using LLMs, with an increasing emphasis on reducing manual schema engineering. Zhang et al.\cite{zhang2024extract} proposed the EDC framework, which decouples extraction, schema definition, and canonicalization, allowing the schema to be either pre-defined or self-generated. Similarly, Ding et al.\cite{ding2024automated} introduced TKGCon, leveraging Wikipedia-derived ontologies, existing theme-specific KGs, and LLM-generated relation sets to build fine-grained and timely theme-specific KGs. SAC-KG~\cite{chen-etal-2024-sac} further integrated generation, verification, and pruning to iteratively construct high-precision domain KGs. In contrast, Su et al.~\cite{su2020automatic} designed a rule-based framework over relational databases for power systems, relying on static schemas and device metadata. Despite these developments, most existing methods either expand pre-existing graphs or apply prompt-based extraction without grounding in authoritative domain templates. 
This gap is particularly evident in the financial domain, where consistency and interpretability are crucial, yet current AKGC approaches seldom incorporate professional, domain-specific schemas.

\subsection{LLMs in Finance}  

The integration of Large Language Models (LLMs) into financial applications has progressed through distinct phases of methodological innovation. Initial efforts focused on domain-specific pretraining, exemplified by Wu et al. \cite{wu2023bloomberggptlargelanguagemodel}, who developed BloombergGPT, a 50-billion-parameter model trained on proprietary financial datasets that demonstrated superior performance in financial NLP tasks. Concurrent work by Yang et al. \cite{yang2023fingptopensourcefinanciallarge} introduced FinGPT, an open-source alternative emphasizing real-time market adaptability through self-supervised learning architectures.
Building on these pioneering efforts, research expanded the scope and capabilities of LLMs in finance. Li et al. \cite{li2024alphafin} proposed Alphafin, a retrieval-augmented stock-chain framework that benchmarks financial analysis by dynamically integrating financial data from multiple sources. Additionally, Yu et al. \cite{yu2024finconsynthesizedllmmultiagent} developed FinCon, a synthesized multi-agent system incorporating conceptual verbal reinforcement to enhance financial decision-making through collaborative agent interactions. 
Recent innovations have targeted multimodal understanding and specialized training regimens. Gan et al. \cite{gan2024mmefinancemultimodalfinancebenchmark} established MME-Finance, a comprehensive benchmark evaluating cross-modal comprehension of financial texts, tables, and charts. 
Building on prior work, recent efforts such as Fin-R1 \cite{liu2025fin} have advanced financial reasoning through high-quality CoT datasets, supervised fine-tuning, and reinforcement learning. The field has progressed toward more sophisticated frameworks incorporating retrieval, multi-agent collaboration, and multimodal reasoning, while ongoing challenges remain in temporal reasoning and explainability for dynamic financial contexts.


\section{Methodology}

\begin{figure*}[ht]
  \centering
  \includegraphics[
    width=0.85\textwidth,
    height=0.35\textheight,
    trim=0pt 0pt 0pt 15pt, 
    clip
  ]{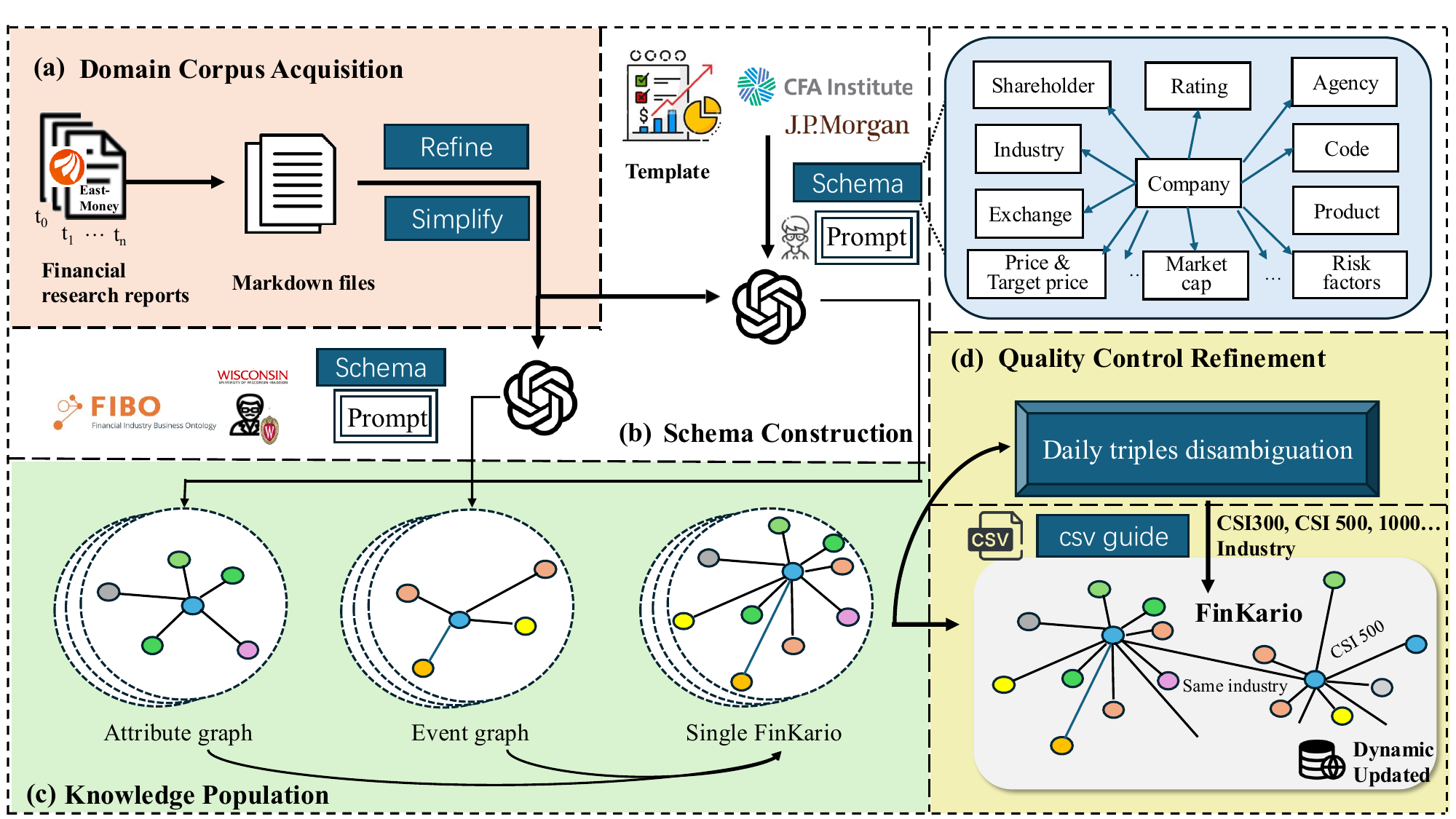}%
  \\[-0.2em]               
  \includegraphics[
    width=0.85\textwidth,
    height=0.33\textheight,
    trim=0pt 40pt 0pt 40pt, 
    clip
  ]{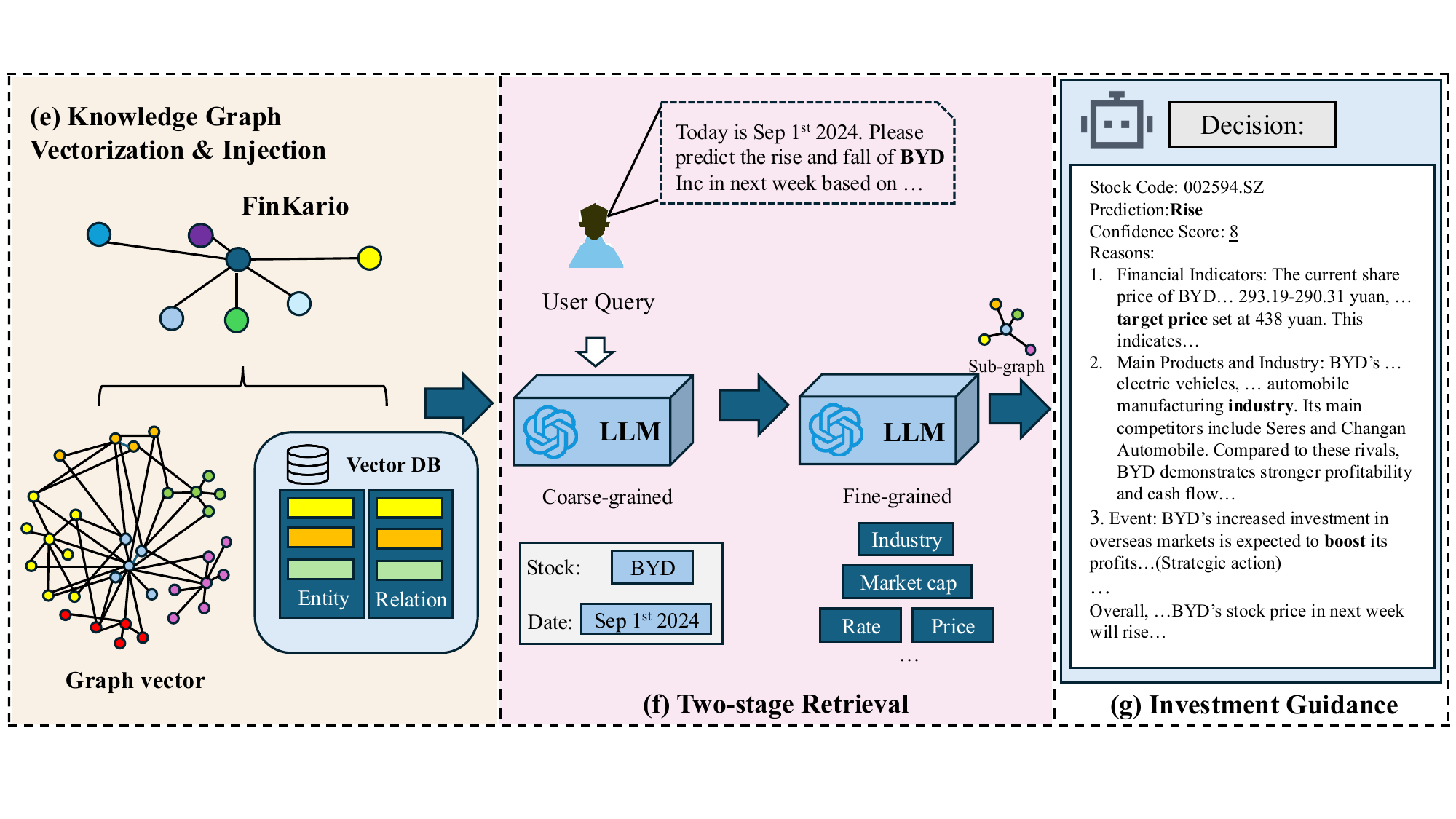}
  \caption{The overall framework of FinKario and FinKario‑RAG: (a)–(d) The construction process of FinKario; (e)–(g) Details of the FinKario‑RAG pipeline.}
  \label{fig:framework}
\end{figure*}

\subsection{FinKario construction}

To support structured interpretation of financial narratives, we introduce a dual-schema design comprising an \textbf{Attribute Graph} and an \textbf{Event Graph}. These are constructed via a schema-guided extraction function:
\[
\mathcal{F}: (\mathcal{D} \times \mathcal{S}) \rightarrow \mathcal{G},
\]
where \(\mathcal{F}\) is a schema-guided extraction function that takes as input a document corpus \(\mathcal{D}\) and a schema \(\mathcal{S}\), and outputs a structured graph \(\mathcal{G}\).

The \underline{Attribute Graph} focuses on relatively stable entity-level properties, such as a firm's industry, exchange, code, product lines, and risk factors. These attributes provide foundational background knowledge and support context-aware retrieval tasks. 
In contrast, the \underline{Event Graph} extends factual information by capturing the underlying drivers of observed financial metrics, such as factors that may trigger a rise in profitability, through a diverse set of \emph{driven categories} (e.g., strategic actions or technological innovation).

Although the Attribute Graph offers a relatively static and straightforward snapshot of the current position and capabilities of a company, the Event Graph provides richer context with greater interpretative flexibility. It allows for deeper insight into a firm's strategic direction, potential growth trajectories, and underlying decision logic, enabling more informed financial analysis and forecasting.

Table~\ref{tab:comparison} presents our knowledge graph alongside existing financial knowledge graphs. Compared to the existing ones, which rely on manually defined schemas and lack comprehensive knowledge updates throughout the process. To construct these graphs in a fully automated manner, we introduce \textbf{FinKario}, a dataset built upon four primary modules, as depicted in Figure~\ref{fig:framework}:
\emph{Domain Corpus Acquisition}, \emph{Schema Construction}, \emph{Knowledge Population}, and \emph{Quality Control Refinement}. 
Each module fulfills a distinct role in automatically transforming raw financial reports into a robust KG without any manually predefined schemas or ontologies. 

\subsubsection{Domain Corpus Acquisition}
We begin by collecting raw financial research reports from the East Money website\footnote{\url{https://www.eastmoney.com}}. To facilitate text parsing, we employ MinerU\footnote{\url{https://github.com/opendatalab/MinerU}} \cite{wang2024mineru}, which converts each report into a standardized Markdown format. We then perform a refinement step to remove non-informative content such as disclaimers, images, and repeated legal statements, ensuring that the resulting corpus \(\mathcal{D'}\) contains only the most relevant textual information for downstream processing.

\begin{figure}[ht]
    \centering
    \includegraphics[width=1.0\linewidth,
    trim=10pt 70pt 0pt 50pt, 
    clip
    ]{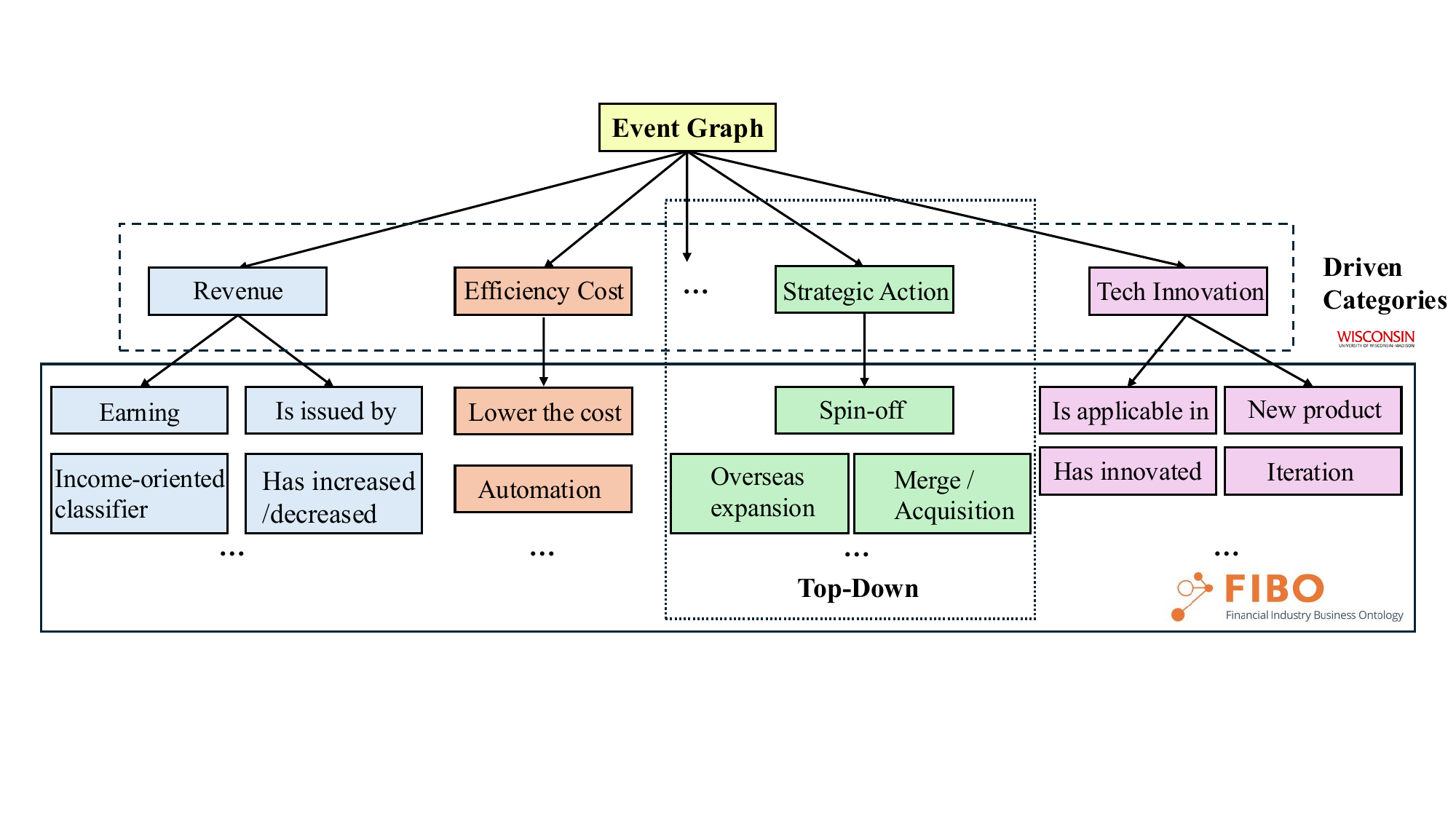}
    \caption{Tree-Structured Schema for Event Graph.}
    \label{fig:event-schema} 
\end{figure}

\subsubsection{Schema Construction}

In this module, we construct two distinct but complementary schemas: the Attribute Graph Schema and the Event Graph Schema.

\noindent $\bullet$ \textbf{Schema for Attribute Graph.}
We leverage standardized equity research templates from authoritative sources (e.g., the CFA Institute\footnote{\url{https://www.cfainstitute.org/sites/default/files/-/media/documents/support/research-challenge/challenge/rc-equity-research-report-essentials.pdf}} and J.P. Morgan\footnote{\url{https://www.wallstreetprep.com/knowledge/sample-equity-research-report/}}) as reference guides. These templates, designated as \(\theta_{\text{CFA}}\) and \(\theta_{\text{JPM}}\), capture the core structure and content of high-quality financial reports.
Based on their organization, we design prompts to guide the LLM in identifying core attribute relation types. Formally, we define:
$\mathcal{S}_A = \text{LLM}(\text{Prompt}_{\text{attr}}; \theta_{\text{CFA}}, \theta_{\text{JPM}})$, and \(\mathcal{S}_A\) is the set of attribute-level relation types such as \emph{Industry}, \emph{Risk Factors}, and \emph{Exchange}. 
The schema \(\mathcal{S}_A\) serves as the foundation for attribute-level knowledge population. Figure~\ref{fig:framework} details the schema in part (b).

\noindent $\bullet$ \textbf{Schema for Event Graph.}
We construct a hierarchical event schema \(\mathcal{S}_E\) using a top-down approach. At the first level, we generate high-level driven categories by prompting the LLM based on the institutional template \(\theta_{\text{WIS}}\) from the University of Wisconsin~\footnote{\url{https://eiexchange.com/content/the-causal-analysis-a-great-way-to-tell-the-financial-story}}:
$\mathcal{C} = \text{LLM}(\text{Prompt}_{\text{cat}}; \theta_{\text{WIS}})$,
where \(\mathcal{C} = \{c_1, c_2, \dots, c_m\}\) represents the set of high-level event categories.
For each category \(c_i \in \mathcal{C}\), we construct prompts grounded in the Financial Industry Business Ontology (FIBO)~\footnote{\url{https://spec.edmcouncil.org/fibo/ontology}}, denoted as \(\mathcal{O}_{\text{FIBO}}\), to generate corresponding low-level event ontology:
$\mathcal{O}_{c_i} = \text{LLM}(\text{Prompt}_{\text{event}}; c_i, \mathcal{O}_{\text{FIBO}})$.
The resulting schema is defined as:
$\mathcal{S}_E = \bigcup_{i=1}^{m} \{(c_i, o) \mid o \in \mathcal{O}_{c_i} \}$.
Figure~\ref{fig:event-schema} visualizes the tree-structured event schema.





\subsubsection{Knowledge Population}

\begin{figure}[t]
  \centering
  \includegraphics[
    width=\linewidth,
    trim=10pt 2pt 5pt 10pt, 
    clip
  ]{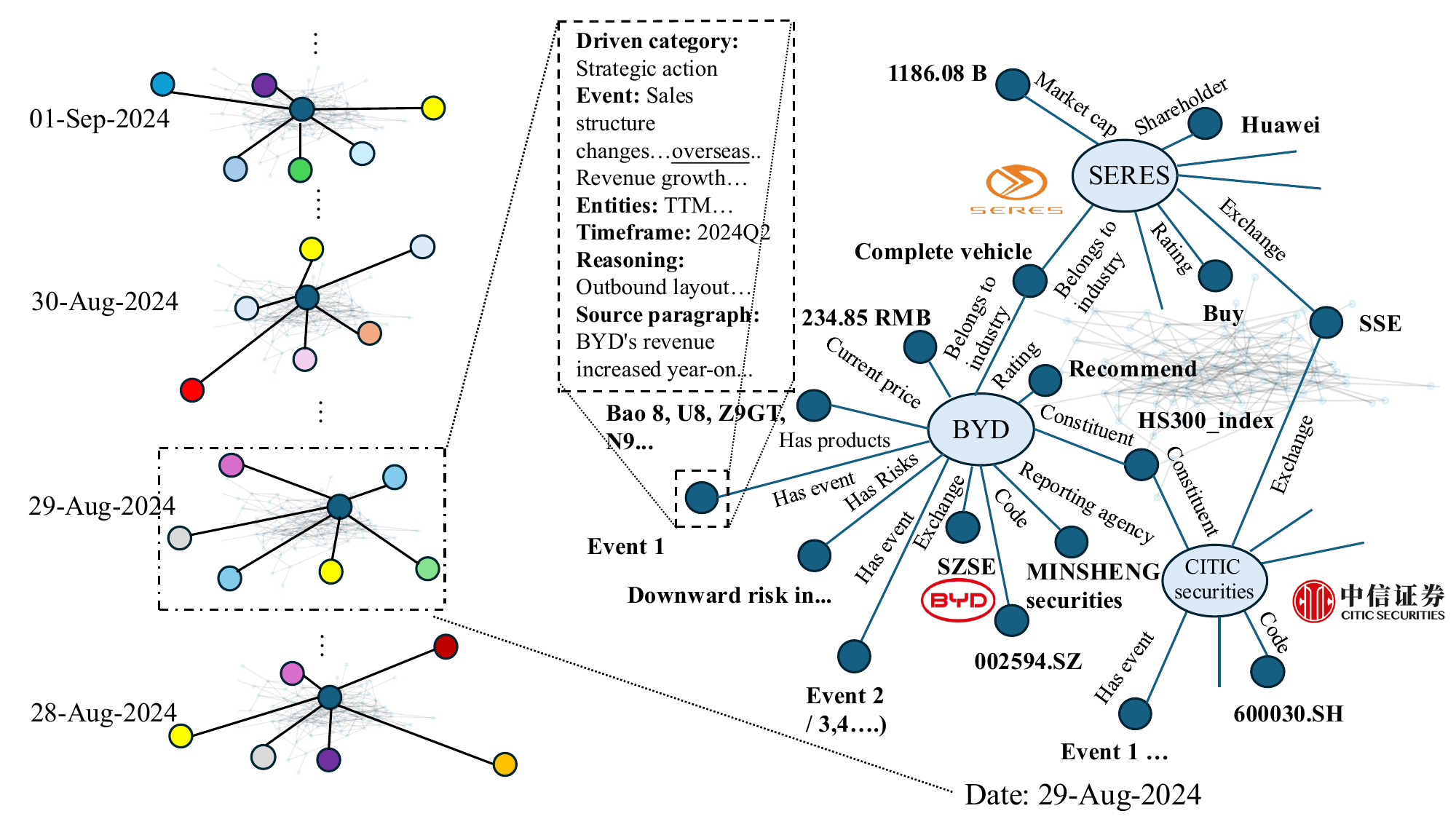}
  \caption{Temporal Visualization of FinKario.}
  \label{fig:visual}
\end{figure}

For each refined Markdown document \(\mathcal{D'}\), entities are extracted at each timestamp \(\tau \in T\) via a dedicated prompt guided by the previously formed schema \(\mathcal{S}_A\) and \(\mathcal{S}_E\):
\[
\mathcal{{E}_A}_\tau; \mathcal{{E}_E}_\tau, \mathcal{{R}_E}_\tau = \text{LLM}(\text{Prompt}, D', \mathcal{S}_A, \mathcal{S}_E, \tau),
\]
where \(\mathcal{{E}_A}_\tau\) and \(\mathcal{{E}_E}_\tau\) denote the set of extracted entities at timestamp \(\tau\).
These timestamped entity sets \(\mathcal{{E}_A}_\tau\) and relation types \(\mathcal{R}_A\) are combined to form the attribute knowledge graph for each stock \(s\):
\[
\mathcal{G}_A^{(s)} = \bigcup_{\tau \in T} \{(e_h, r, e_t, \tau) \mid e_h, e_t \in \mathcal{{E}_A}_\tau, r \in \mathcal{R}_A\},
\]
where \(e_h\) and \(e_t\) denote head and tail entities, respectively.
Subsequently, a dynamic Event Knowledge Graph is constructed separately, guided by a distinct set of schema \(\mathcal{S}_E\):
\[
\mathcal{G}_E^{(s)} = \bigcup_{\tau \in T} \{(e_s, r', e_o, \tau) \mid e_s, e_o \in \mathcal{{E}_E}_\tau, r' \in \mathcal{{R}_E}_\tau\},
\]
where \(e_s\) is the subject entity, \(e_o\) is the object entity, and \(r'\) represents  the trigger relationship \(\mathcal{{R}_E}_\tau\).
Finally, the comprehensive financial knowledge graph for stock \(s\) integrates both attribute and event knowledge graphs:
\[
\mathcal{G}_{\text{FinKario}}^{(s)} = \mathcal{G}_A^{(s)} \cup \mathcal{G}_E^{(s)}.
\]
This integrated graph \(\mathcal{G}_{\text{FinKario}}^{(s)}\) effectively captures structured financial information along with inferred event interactions. Figure~\ref{fig:visual} illustrates the daily temporal structure of FinKario.


\subsubsection{Quality Control Refinement}

To ensure the reliability of the constructed knowledge graph, we implement a refinement module that addresses common issues in financial text extraction, including entity ambiguity, missing numeric values, and extraction errors. Specifically, the module performs entity normalization, attribute completion via the Tushare platform\footnote{\url{https://tushare.pro}}, and error or placeholder correction via the LLM. The complete refinement pipeline is detailed in Algorithm~\ref{alg:quality-refinement}.

\begin{algorithm}[h]
\caption{Quality Control Refinement}
\label{alg:quality-refinement}
\begin{algorithmic}[1]
\Require Raw knowledge graph $\mathcal{G}_{\text{FinKario}}^{(s)}$, reference dictionary $\mathcal{T}_{\text{ref}}$, and LLM(\(\cdot\)) instantiated as GPT-4o-mini
\Ensure Refined knowledge graph $\mathcal{G}_{\text{FinKario}}^{\prime (s)}$

\vspace{0.5em}
\Statex \textbf{// Step 1: Entity Normalization}
\ForAll{entity $e \in \mathcal{G}_{\text{FinKario}}^{(s)}$}
    \If{$e$ is a name variant (e.g., "BYD Inc.", "BYD Auto")}
        \State Replace $e$ with canonical form (e.g., "BYD")
    \EndIf
\EndFor

\vspace{0.5em}
\Statex \textbf{// Step 2: Attribute Completion}
\ForAll{triple $(e_h, r, e_t) \in \mathcal{G}_{\text{FinKario}}^{(s)}$}
    \If{$r$ is a numeric attribute (e.g., "Price", "Cap") and ($e_t$ is missing or lacks unit)}
        \State Query $\mathcal{T}_{\text{ref}}$ for value and unit (e.g., CNY, USD, billions)
        \State Replace $e_t$ with correct value and unit
    \EndIf
\EndFor

\vspace{0.5em}
\Statex \textbf{// Step 3: Error Correction via LLM}
\ForAll{triple $(e_h, r, e_t) \in \mathcal{G}_{\text{FinKario}}^{(s)}$}
    \If{$e_t$ contains a placeholder (e.g., "No relevant information was found", "Extraction error")}
        \State Re-feed the source Markdown passage to \texttt{LLM}(\(\cdot\))
        \State Replace $e_t$ with the corrected output from the LLM
    \EndIf
\EndFor

\State \Return $\mathcal{G}_{\text{FinKario}}^{\prime (s)}$
\end{algorithmic}
\end{algorithm}

\subsection{FinKario-RAG}

The two-stage graph-based retrieval augmented generation pipeline (FinKario-RAG) converts $\mathcal{G}_{\text{FinKario}}^{\prime(s)}=(\mathcal{E},\mathcal{R})$ into actionable investment advice through three interlocking modules, as illustrated in Fig.~\ref{fig:framework}.


\subsubsection{Knowledge Graph Vectorization \& Ingestion}

To support semantic retrieval, the event-augmented financial knowledge graph $\mathcal{G}_{\text{FinKario}}^{\prime(s)}$ is vectorized into three components: entity-level, relation-level, and graph-level representations.

\noindent$\bullet$~\textbf{Entity and Relation Embedding.}  
We encode all entities and relations using a graph encoder $\Phi$ to obtain the structural embedding set:
\[
\mathbf{Z}_{\text{local}} = \Phi(\mathcal{G}_{\text{FinKario}}^{\prime(s)}) = \{ \mathbf{e}_i \}_{i=1}^{|\mathcal{E}|} \cup \{ \mathbf{r}_j \}_{j=1}^{|\mathcal{R}|},
\]
where $\mathbf{e}_i$ and $\mathbf{r}_j$ denote the latent vectors of the $i$-th entity and $j$-th relation, respectively.

\noindent$\bullet$~\textbf{Graph-level Embedding.}  
To capture global context and topological semantics, we also compute a graph-level vector using a readout function \( \rho \), where $\mathbf{g}_{\text{global}} = \rho(\mathcal{G}_{\text{FinKario}}^{\prime(s)})$.

\noindent$\bullet$~\textbf{Vector Store Indexing.}  
The unified representation
$\mathbf{Z}_{\text{FinKario}} = \mathbf{Z}_{\text{local}} \cup \{\mathbf{g}_{\text{global}}\}$
is normalized and stored in the vector database \( \mathbb{V} \), which supports efficient maximum inner product search during downstream retrieval.

\subsubsection{Two-stage Retrieval}

Given a user query $q$, the system first encodes it into a dense vector $\mathbf{h}_q = \Psi(q)$ via a language model encoder $\Psi$. Retrieval is performed in two stages:

\noindent$\bullet$ \textbf{Coarse-grained Retrieval.}  
The first stage aims to identify rough semantic anchors such as relevant stocks and dates. This is achieved by matching $\mathbf{h}_q$ against indexed stock and date representations in the vector store:
\[
\mathbb{V}_{\text{coarse}} = \mathcal{R}_{\text{coarse}}(\mathbf{h}_q, \mathbb{V}, k_c),
\]
returning the top-$k_c$ coarse candidates (e.g., \texttt{BYD}, Sep 1\textsuperscript{st} 2024).

\noindent$\bullet$ \textbf{Fine-grained Retrieval.}  
Building on the coarse results, a finer-grained retrieval is conducted to collect surrounding financial entities—such as \texttt{industry}, \texttt{market cap}, and \texttt{price} by searching over relevant portions of the vector set:
\[
\mathbb{V}_{\text{fine}} = \mathcal{R}_{\text{fine}}(\mathbf{h}_q, \mathbb{V}_{\text{coarse}}, k_f),
\]
where \(k_f\) represents the number of related entities from fine-grained process.
To facilitate structured reasoning, the retrieved fine-level vectors are mapped back to their original graph context to reconstruct a semantically aligned subgraph:
\[
\mathcal{G}_{\text{sub}} = \texttt{Mapping}(\mathbb{V}_{\text{fine}}), \quad \mathcal{G}_{\text{sub}} \subseteq \mathcal{G}_{\text{FinKario}}^{\prime(s)}.
\]
where \texttt{Mapping}(\(\cdot\)) refers to a lookup procedure that aligns vector-retrieved entities with their corresponding nodes and edges in the original graph \(\mathcal{G}_{\text{FinKario}}^{\prime(s)}\).
This two-stage process allows FinKario-RAG to retrieve a compact, semantically coherent subgraph for downstream reasoning, preserving both high-level user intent and local financial context.

\subsubsection{Investment Guidance}

The subgraph $\mathcal{G}_{\text{sub}}$ and the user query $q$ are jointly fed into the final reasoning model:
\[
y = \text{LLM}_{\text{Analyst}}(q, \mathcal{G}_{\text{sub}}),
\]
where $y$ includes a predicted movement label (e.g., \texttt{Rise} or \texttt{Fall}), an associated confidence level, and a textual rationale grounded in the retrieved knowledge.
This completes the FinKario-RAG pipeline by converting graph-derived evidence into interpretable and actionable investment guidance.

\section{Experiment Results}

\subsection{Experiment Setup}
Given a universe of stocks $\mathcal{S}$, for any stock $s \in \mathcal{S}$ on a given trading day $t$, 
we evaluate a long-only trading strategy driven by FinKario-RAG signals. The strategy operates as follows:
(1) \textbf{Signal Generation.} On trading day $t$, FinKario-RAG generates a signal $\gamma_{s,t}$ for stock $s$; (2) \textbf{Entry Rule.} If $\gamma_{s,t}$ indicates a buy signal, we initiate a position by purchasing the stock at the closing price $c_{s,t+1}$ on day $t+1$; (3) \textbf{Exit Rule.} The position is held until the last trading day of the following week, denoted as $\tau(t)$, , at which point the stock is sold at the closing price $c_{s,\tau(t)}$.


\subsection{Dataset \& Metrics}
We evaluate our model using a multi-source dataset that combines textual and financial data. The raw research reports are collected from the East Money website, covering the period from 2024-08-28 to 2025-02-28. Corresponding stock price data for backtesting is obtained from Tushare, spanning from 2024-08-28 to 2025-03-07. In addition, we incorporate index components and industry classification information provided by Wind platform to support graph construction and semantic enrichment. 

To evaluate the performance of our model, we adopt six widely used metrics: \textbf{Annualized Rate of Return (ARR)} measures the compound annual growth rate of the portfolio value over the evaluation period. \textbf{Volatility (VOL)} quantifies the annualized standard deviation of weekly returns, indicating the portfolio's risk level. \textbf{Sharpe Ratio (SR)} quantifies risk-adjusted performance by dividing the annualized return by the annualized volatility. \textbf{Maximum Drawdown (MDD)} measures the largest peak-to-trough decline in portfolio value, representing the worst-case loss scenario. \textbf{Calmar Ratio (CR)} assesses risk-adjusted returns by dividing the annualized return by the absolute maximum drawdown. 
\textbf{Accuracy (ACC)} measures the percentage of correct directional predictions generated by FinKario-RAG trading signals.
Together, these metrics provide a comprehensive view of both predictive quality and practical investment performance of our model.

\subsection{Baseline}
Our proposed approach is evaluated against four categories of baselines: 

\noindent$\bullet$~\textbf{Market Indices.} Market indices are standard passive benchmarks. We report results on several representative indices, including the \textbf{CSI 300}, \textbf{CSI 500}, \textbf{SSE Composite Index}, and \textbf{SSE Dividend Index}, which cover major segments of the Chinese market.

\noindent$\bullet$~\textbf{Vanilla LLMs.} We include general-purpose language models such as Qwen3-8B \cite{yang2025qwen3} and GPT-4o-mini \cite{hurst2024gpt}, 
which are not specifically tuned for financial tasks.

\noindent$\bullet$~\textbf{Financial Domain LLMs.} We evaluate several open-source financial language models, including FinMA \cite{xie2023pixiu}, FinGPT \cite{yang2023fingptopensourcefinanciallarge}, DISC-FinLLM \cite{chen2023disc}, XuanYuan-6B \footnote{https://huggingface.co/Duxiaoman-DI/XuanYuan-6B} and Stock-Chain \cite{li2024alphafin}. 
These models are tailored for financial forecasting and investment recommendations, making them suitable for downstream backtesting and comparison.

\noindent$\bullet$~\textbf{Financial Institutions.} 
To the best of our knowledge, this is the first work to incorporate real-world institutional strategies as a baseline. The selected institutions—Tianfeng, Southwest, Sinolink, Soochow, Guolian-Mingsheng, Guosen, Huaan, Kaiyuan, China Fortune, and China Post—were chosen as leading brokerages that frequently publish research reports, each appearing in at least 300 reports.

\begin{figure*}[ht] 
    \centering
    \includegraphics[width=0.9\textwidth,
    trim=0pt 15pt 0pt 10pt, 
    clip]{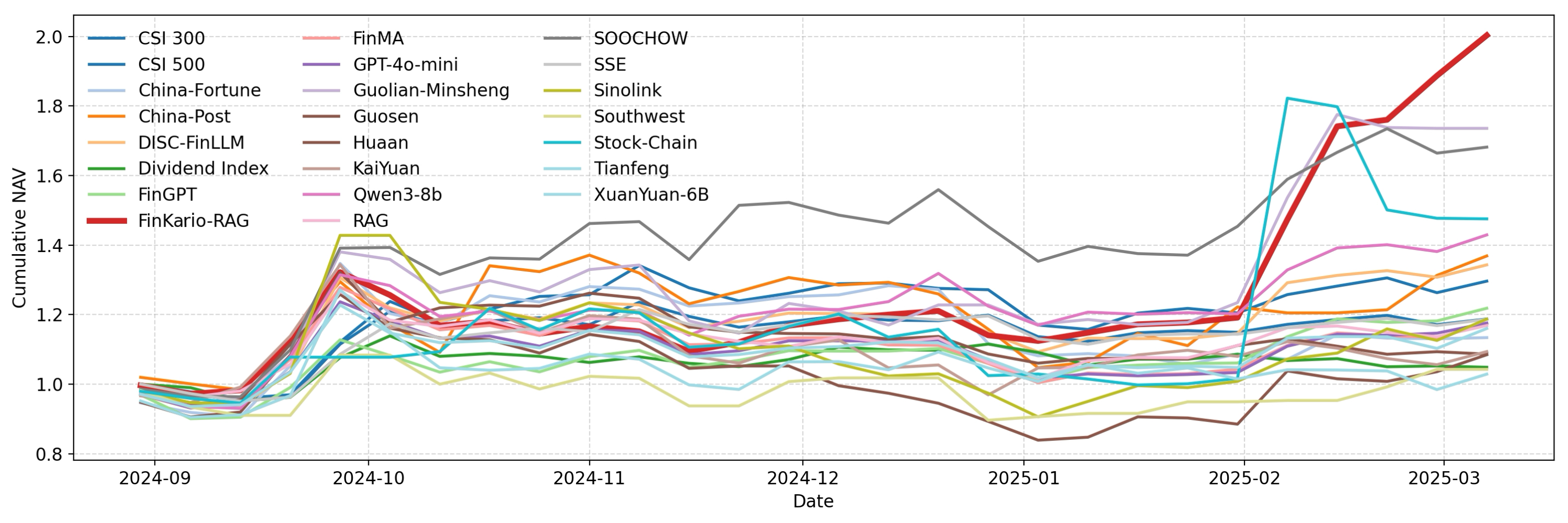} 
    \caption{{Accumulated returns (AR) of each baseline strategy on the financial‐report dataset from August 28, 2024 to March 7, 2025. The figure shows the net asset value (NAV) curves over the weekly backtesting period.}}
    \label{fig:nav_curve} 
    
\end{figure*}


\begin{table*}[htbp]
\centering
\caption{Performance comparison across market indices, vanilla LLMs, financial domain LLMs, and institutional strategies.}
\label{tab:model-results}
\renewcommand{\arraystretch}{0.8}
\setlength{\tabcolsep}{16pt}
\begin{tabular}{lrrrrrc}
\toprule
\textbf{Model} & \textbf{ARR↑} & \textbf{VOL↓} & \textbf{SR↑} & \textbf{MDD↓} & \textbf{CR↑} & \textbf{ACC↑} \\
\midrule
CSI 300         & 0.392 & 0.295 & 1.330 & 0.091 & 4.332 & - \\
CSI 500         & 0.648 & 0.342 & 1.894 & 0.137 & 4.729 & - \\
SSE             & 0.390 & 0.265 & 1.471 & \cellcolor[rgb]{ .98, .635, .643}0.082 & 4.748 & - \\
Dividend Index  & 0.096 & \cellcolor[rgb]{ .973, .412, .42}0.233 & 0.411 & \cellcolor[rgb]{ .973, .412, .42}0.079 & 1.208 & - \\
\midrule\midrule
Qwen3-8b        & 0.941 & 0.459 & 2.051 & 0.132 & 7.130 & 0.475 \\
GPT-4o-mini     & 0.351 & 0.372 & 0.944 & 0.178 & 1.977 & 0.471 \\
RAG (4o-mini)   & 0.336 & 0.360 & 0.932 & 0.197 & 1.703 & 0.559 \\
\midrule\midrule
FinMA           & 0.348 & 0.389 & 0.895 & 0.214 & 1.623 & 0.479 \\
FinGPT          & 0.443 & 0.327 & 1.355 & \cellcolor[rgb]{ .988, .89, .902}0.103 & 4.294 & 0.475 \\
DISC-FinLLM     & 0.729 & 0.468 & 1.559 & 0.163 & 4.462 & 0.474 \\
XuanYuan-6B     & 0.318 & 0.373 & 0.852 & 0.170 & 1.868 & 0.471 \\
Stock-Chain     & 1.177 & 1.211 & 0.971 & 0.190 & 6.182 & 0.546 \\
\midrule\midrule
Tianfeng            & 0.054 & 0.542 & 0.100 & 0.225 & 0.242 & 0.411 \\
Southwest           & 0.121 & 0.485 & 0.249 & 0.173 & 0.701 & 0.492 \\
Sinolink            & 0.391 & 0.648 & 0.604 & 0.365 & 1.070 & 0.438 \\
SOOCHOW             & \cellcolor[rgb]{ .988, .89, .902}1.625 & 0.522 & \cellcolor[rgb]{ .98, .635, .643}3.115 & 0.132 & \cellcolor[rgb]{ .98, .635, .643}12.311 & 0.557 \\
Guolian-Minsheng    & \cellcolor[rgb]{ .98, .635, .643}2.012 & 0.647 & \cellcolor[rgb]{ .988, .89, .902}3.108 & 0.169 & \cellcolor[rgb]{ .988, .89, .902}11.880 & \cellcolor[rgb]{ .98, .635, .643}0.575 \\
Guosen              & 0.167 & \cellcolor[rgb]{ .988, .89, .902}0.456 & 0.366 & 0.197 & 0.845 & 0.460 \\
Huaan               & 0.170 & 0.471 & 0.361 & 0.333 & 0.509 & 0.435 \\
KaiYuan             & 0.181 & 0.473 & 0.383 & 0.279 & 0.650 & 0.552 \\
China-Fortune       & 0.263 & 0.537 & 0.489 & 0.216 & 1.218 & \cellcolor[rgb]{ .988, .89, .902}0.573 \\
China-Post          & 0.830 & 0.559 & 1.485 & 0.236 & 3.519 & 0.440 \\
\midrule\midrule
\textbf{FinKario-RAG} & \cellcolor[rgb]{ .973, .412, .42}2.633 & 0.534 & \cellcolor[rgb]{ .973, .412, .42}4.926 & 0.172 & \cellcolor[rgb]{ .973, .412, .42}15.315 & \cellcolor[rgb]{ .973, .412, .42}0.581 \\
\bottomrule
\end{tabular}
\end{table*}

\subsection{Experimental Results}

Figure \ref{fig:nav_curve} illustrates that FinKario-RAG consistently outperforms all benchmark models in cumulative returns, showcasing the effectiveness of our knowledge graph-enhanced retrieval framework. In late September 2024, most strategies surged in response to favorable Chinese fiscal policies, followed by a period of pullback and sideways movement, during which SOOCHOW's strategy remained notably stable. 

A major turning point occurred in early February 2025, 
as many strategies rebounded.
Stock-Chain exhibited a sharp spike, while FinKario-RAG entered a phase of steady and accelerating growth, ultimately surpassing nearly all competitors by early March. These results highlight FinKario-RAG's adaptability to market shifts, effectiveness in capturing transient signals, and robust performance across volatile periods through its structured graph-based design.

In addition to the visual insights from cumulative NAV trends, Table~\ref{tab:model-results} presents a quantitative comparison that further validates FinKario-RAG’s superiority. FinKario-RAG achieves the highest scores in ARR (2.633), Sharpe Ratio (4.926), Calmar Ratio (15.315), and ACC (0.581), while maintaining a moderate Maximum Drawdown (MDD) of 0.172. These results reflect a balanced and effective investment strategy.

In terms of ARR, FinKario-RAG outperforms Guolian-Minsheng (2.012) by 30.8\%, SOOCHOW (1.625) by 62.0\%, and exceeds Stock-Chain by a significant 123.7\%. For risk-adjusted returns, FinKario-RAG’s SR and CR represent improvements of 58.1\% and 24.4\% over the runner-up performers. Although its volatility (0.534) is not the lowest, it remains well-controlled relative to its high returns.

Regarding predictive accuracy, FinKario-RAG achieves a leading ACC of 0.581, outperforming Guolian-Minsheng (0.575), China Fortune (0.573), and RAG (0.559). While some institutional strategies also demonstrate strong accuracy, 
performance varies considerably across institutions, with the maximum accuracy gap reaching 0.164.

Overall, FinKario-RAG strikes a robust risk-reward balance. Its knowledge graph-enhanced design delivers 
both superior profitability and effective risk control, positioning it as a state-of-the-art approach 
for 
LLM-based quantitative investment.


\begin{table}[t]
  \centering
  \caption{Ablation study on the impact of different knowledge sources. $'w/'$ uses other knowledge sources instead of FinKario; $'w/o'$ removes parts of FinKario for ablation.}
  \label{tab:ablation_study_knowledge}
  \renewcommand{\arraystretch}{0.9}
  \setlength{\tabcolsep}{1.5pt}
  \scalebox{1.0}{
\begin{tabular}{l|l|cccc}
\toprule
                   & \textbf{Knowledge}       & \textbf{ARR↑} & \textbf{SR↑} & \textbf{MDD↓} & \textbf{ACC↑}\\
\midrule
\multirow{4}{*}{FinKario-RAG} 
                   & w/ Research report        & 0.336         & 0.932        & 0.197    & 0.559     \\
                   & w/ HiDy & 0.462 & 1.353 & 0.174 & 0.455\\
                   \cmidrule(lr){2-6}
                   & w/o Event graph      & 0.386         & 0.903        & 0.177 & 0.474        \\
                   & w/o Attribute graph      & 2.230         & 4.691        & 0.181   & 0.433    \\
                   \cmidrule(lr){2-6}
                   & \textbf{FinKario (Ours)} & \textbf{2.633}& \textbf{4.926}& \textbf{0.172}  & \textbf{0.581} \\
\bottomrule
\end{tabular}
}
\end{table}

\begin{table}[t]
  \centering
  \caption{Ablation study of varied retrieval approach.}
  \label{tab:ablation_study_method}
  \renewcommand{\arraystretch}{0.9}
  \setlength{\tabcolsep}{2pt}
  \scalebox{1.0}{
\begin{tabular}{l|l|cccc}
\toprule
                   & \textbf{Method}       & \textbf{ARR↑} & \textbf{SR↑} & \textbf{MDD↓} & \textbf{ACC↑}\\
\midrule
\multirow{3}{*}{FinKario} 
                   & Vanilla RAG            & 0.377         & 0.758        & \textbf{0.120}     & 0.413    \\
                   & LightRAG      & 0.821         & 1.313        & 0.140  & 0.495       \\
                   & \textbf{FinKario-RAG (Ours)} & \textbf{2.633}& \textbf{4.926}&  0.172  & \textbf{0.581}\\
\bottomrule
\end{tabular}
}
\end{table}

\subsection{Ablation Study}

We conduct two ablation studies. All variants are built on the same backbone, GPT-4o-mini, serving as both the chat-based reasoning module and the embedding encoder for vector retrieval, ensuring consistency across comparisons. 

\subsubsection{Varied Knowledge Source Injection}

Table~\ref{tab:ablation_study_knowledge} presents an ablation study quantifying the significance of varied knowledge sources. Removing the Event graph (w/o Event graph) results in a dramatic decrease in ARR, showing an 87.2\% drop, from 2.633 to 0.336, and a reduction in SR by 81.1\%, from 4.926 to 0.932. In contrast, removing the Attribute graph (w/o Attribute graph) also leads to performance degradation, but the decline is less pronounced across all indicators, underscoring the complementary role of the Event graph in enhancing model performance. Additionally, we examine the impact of raw markdown content, which leads to a noticeable decline in performance. The model struggles to filter out useful information, particularly when processing long-text, highlighting its inefficiency in handling unstructured data. Furthermore, we compare the open-source knowledge base HiDy~\footnote{\url{https://zenodo.org/records/12630355}}, which integrates multiple sources of financial data. The results suggest that HiDy provides valuable supplementary knowledge but does not outperform the full knowledge graph structure in improving performance, with a reduction in SR by 72.5\%.

\subsubsection{Varied Retrieval Approach}

Table~\ref{tab:ablation_study_method} highlights the ablation study on varied retrieval approaches. We integrate FinKario with traditional retrieval methods. The Vanilla RAG approach shows lowest performance across all metrics, with ARR and SR both dropping by 85.7\% and 80.5\%, respectively. In contrast, the LightRAG method shows modest gains, with ARR and SR improvements of 23.5\% and 17.5\%, respectively, suggesting the effectiveness of the graph-based retrieval approach. However, it still underperforms compared to FinKario-RAG. These results underscore the advantage of FinKario-RAG framework, which delivers superior performance across all metrics.







\begin{figure}[t] 
    \centering
    \vspace{-6px}
    \subfigure[FinKario-RAG]
    {\includegraphics[width=0.232\textwidth,
    trim=5pt 10pt 5pt 5pt, 
    clip]{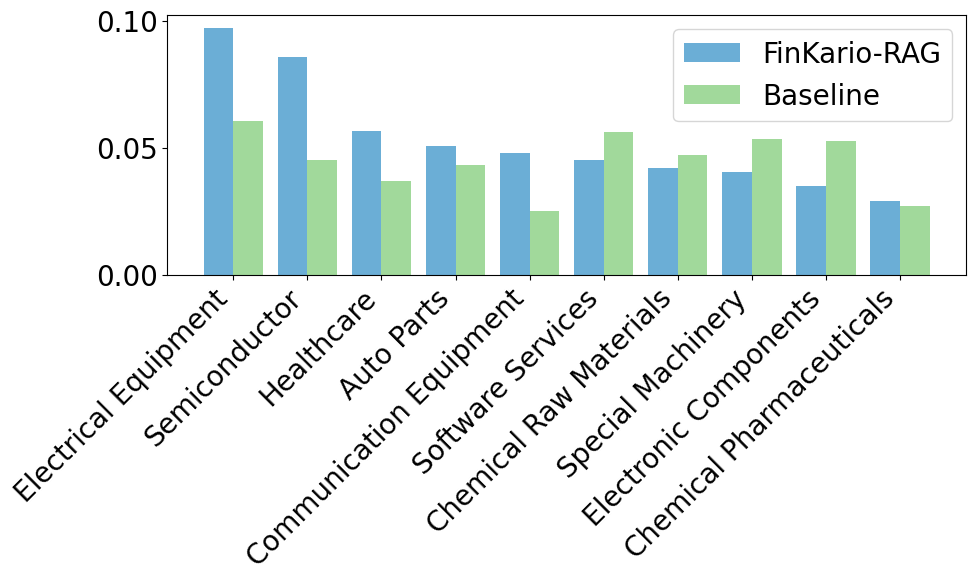}
    \label{subfig:FinKario}}
    \subfigure[Qwen3-8b]
    {\includegraphics[width=0.232\textwidth,
    trim=5pt 10pt 5pt 5pt, 
    clip]{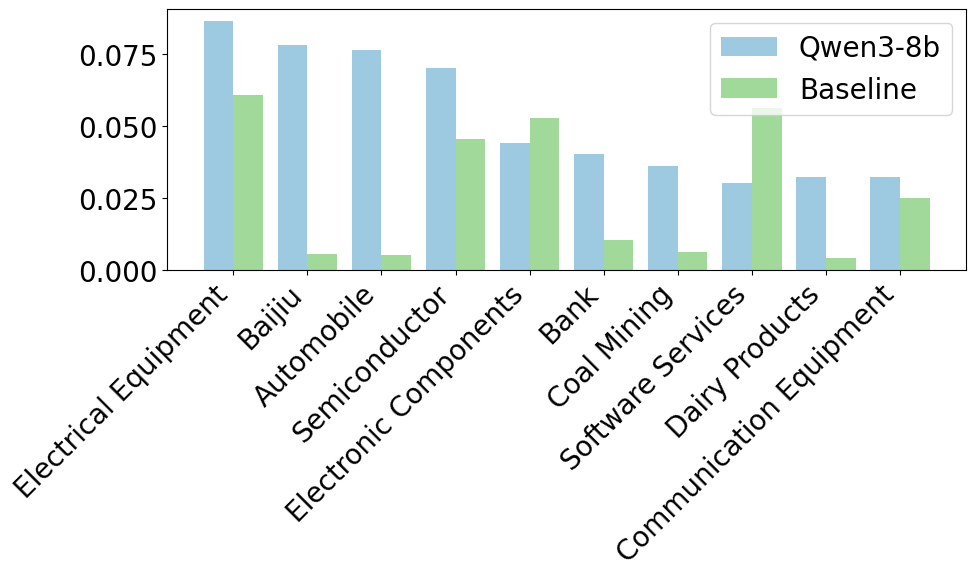}\label{subfig:qwen3}}
    \\[-0.10em] 
    \subfigure[Guolian-Mingsheng]
    {\includegraphics[width=0.232\textwidth,
    trim=5pt 10pt 5pt 5pt, 
    clip]{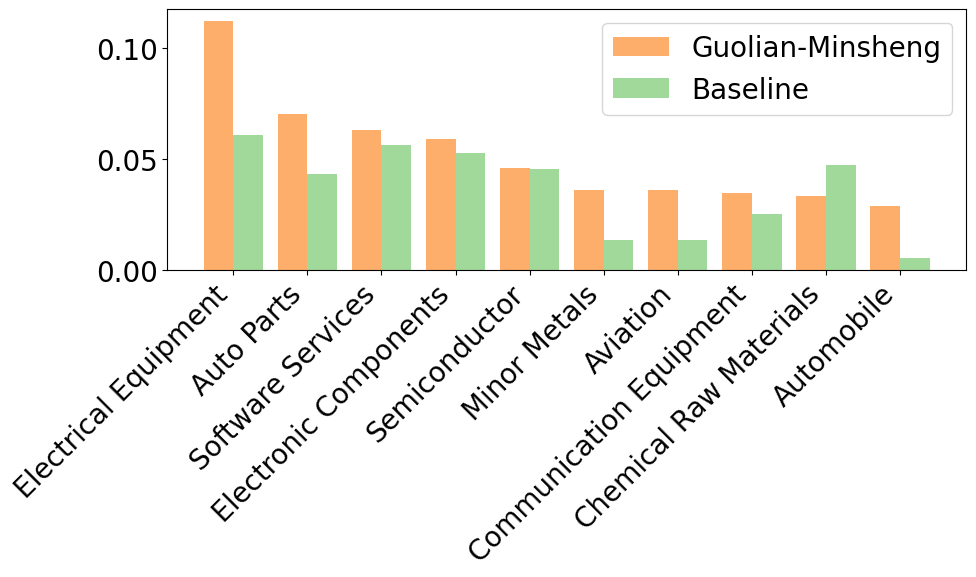}\label{subfig:guolian}}
    \subfigure[Stock-Chain]
     {\includegraphics[width=0.232\textwidth,
     trim=5pt 10pt 5pt 5pt, 
     clip]{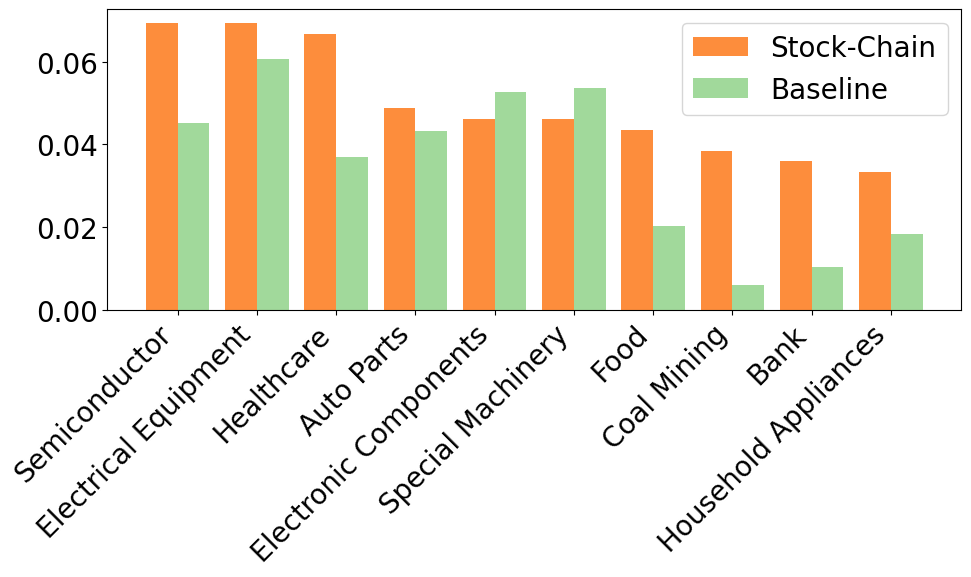}\label{subfig:stock-chain}}
    \vspace{-5px}
    \caption{Visualization of Model Industry Preferences vs.  Baseline Preferences. The baseline reflects the original industry distribution derived from raw research reports.}
    \vspace{-5px}
    \label{fig:Industry-preference} 
\end{figure}

\subsection{Case Study}

To illustrate model behavior, we compare FinKario-RAG with the best-performing models in each category: Qwen3-8B (vanilla LLM), Guolian-Mingsheng (institution), and Stock-Chain (financial LLM). As depicted in Figure~\ref{fig:Industry-preference}, FinKario-RAG exhibits a strong concentration in high-growth sectors such as Electrical Equipment, Semiconductor, and Healthcare—an allocation strategy that closely aligns with the technology-led market rally observed around February 2025.

In contrast, Qwen3-8B and Stock-Chain present broader and less focused industry allocations across their top 3–4 sectors. Guolian-Mingsheng narrows its focus mainly to Auto Parts and Semiconductor, which aligns with FinKario-RAG to some extent and helps sustain performance in the later market stages. Although Stock-Chain also overweights Semiconductor, its simultaneous heavy allocation to sectors like Food, Coal Mining and Bank results in an implicit internal hedging effect, partially diluting return potential during sector rallies. FinKario-RAG’s industry targeting strategy, by contrast, reflects higher consistency and adaptability to sector momentum.

\section{Conclusion}

This work presents \textbf{FinKario}, a fully automated dataset constructed with event-enhanced knowledge grounded in professional institutional templates, ensuring both domain alignment and scalable dynamic updates. 
By integrating \textbf{FinKario-RAG}, a two-stage graph-based RAG mechanism, our approach overcomes the limitations of single-target retrieval and reduces hallucination risk when handling large, dynamically evolving graphs.
Extensive experiments demonstrate that FinKario-RAG significantly outperforms both vanilla LLMs and state-of-the-art FinLLMs in stock trend forecasting. To the best of our knowledge, this is the first benchmark against real-world institutional strategies, providing a more grounded and practical performance comparison. Looking forward, FinKario can be extended to incorporate multi-modal financial inputs, such as tables, charts, and time-series data from research reports, to further enrich its retrieval context and enhance predictive robustness.


\appendix
\section{Prompts in FinKario}

\subsection{The Prompt for Acquiring Schema of Attribute Graph}
\begin{figure*}[ht]
  \centering
  \includegraphics[
    width=\textwidth,
    trim=10pt 2pt 5pt 10pt, 
    clip
  ]{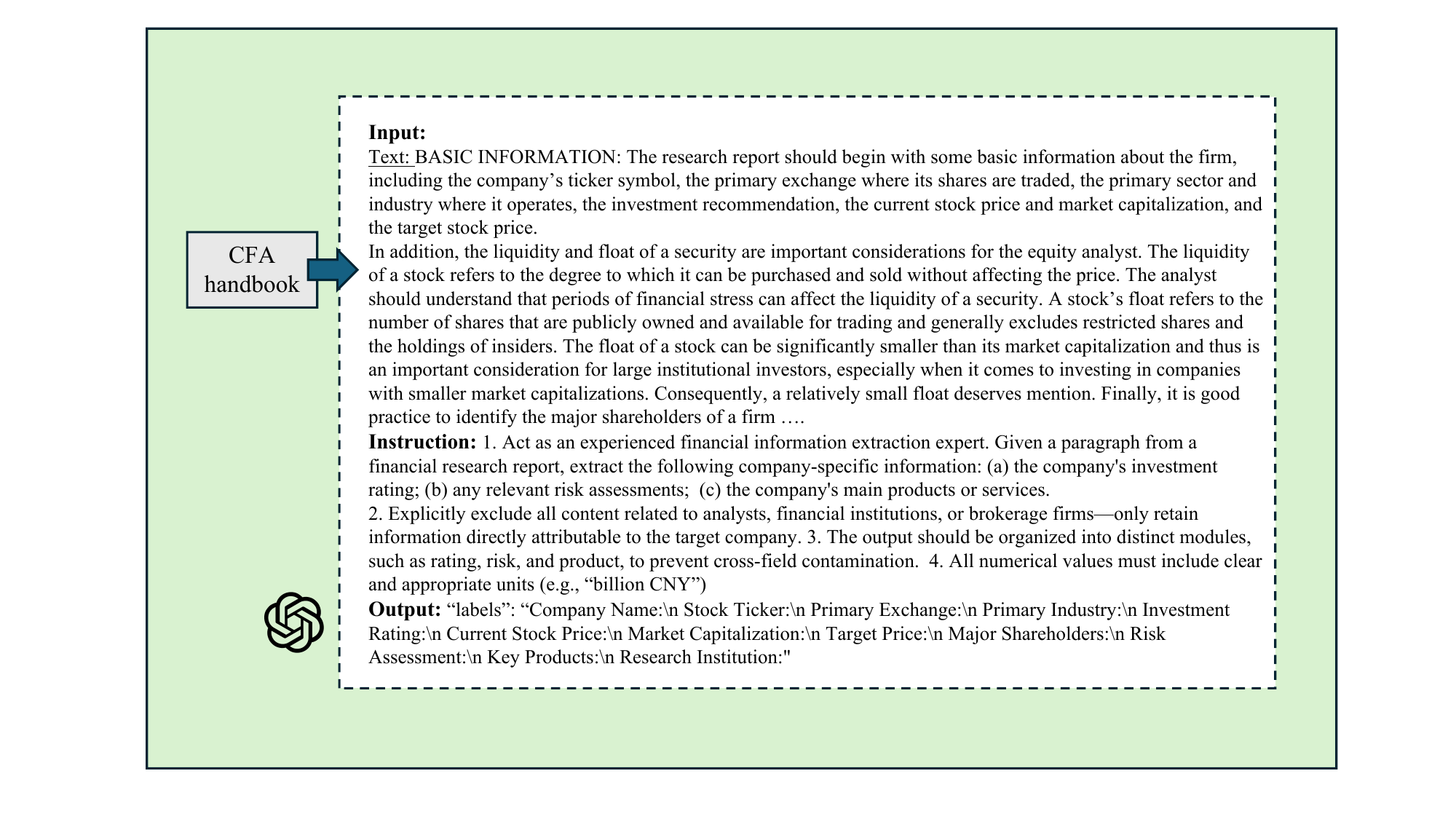}
  \caption{Prompt for acquiring schema of Attribute graph.}
  \label{fig:schema_attribute}
\end{figure*}

We leverage standardized equity research templates from authoritative sources (e.g., the CFA Institute and J.P. Morgan) as reference guides. We design a prompt that guides the model to identify core company attributes for schema construction, including name, ticker, rating, market capitalization, and more. Specifically, the attribute schema comprises 11 relation types:
Stock Ticker, Primary Exchange, Primary Industry, Investment Rating, Current Stock Price, Market Capitalization, Target Price, Major Shareholders, Risk Assessment, Key Products, Research Institution.
Figure~\ref{fig:schema_attribute} illustrates an example of the automatically acquired schema for our Attribute graph.


\begin{figure*}[ht]
  \centering
  \includegraphics[
    width=\textwidth,
    trim=10pt 2pt 5pt 10pt, 
    clip
  ]{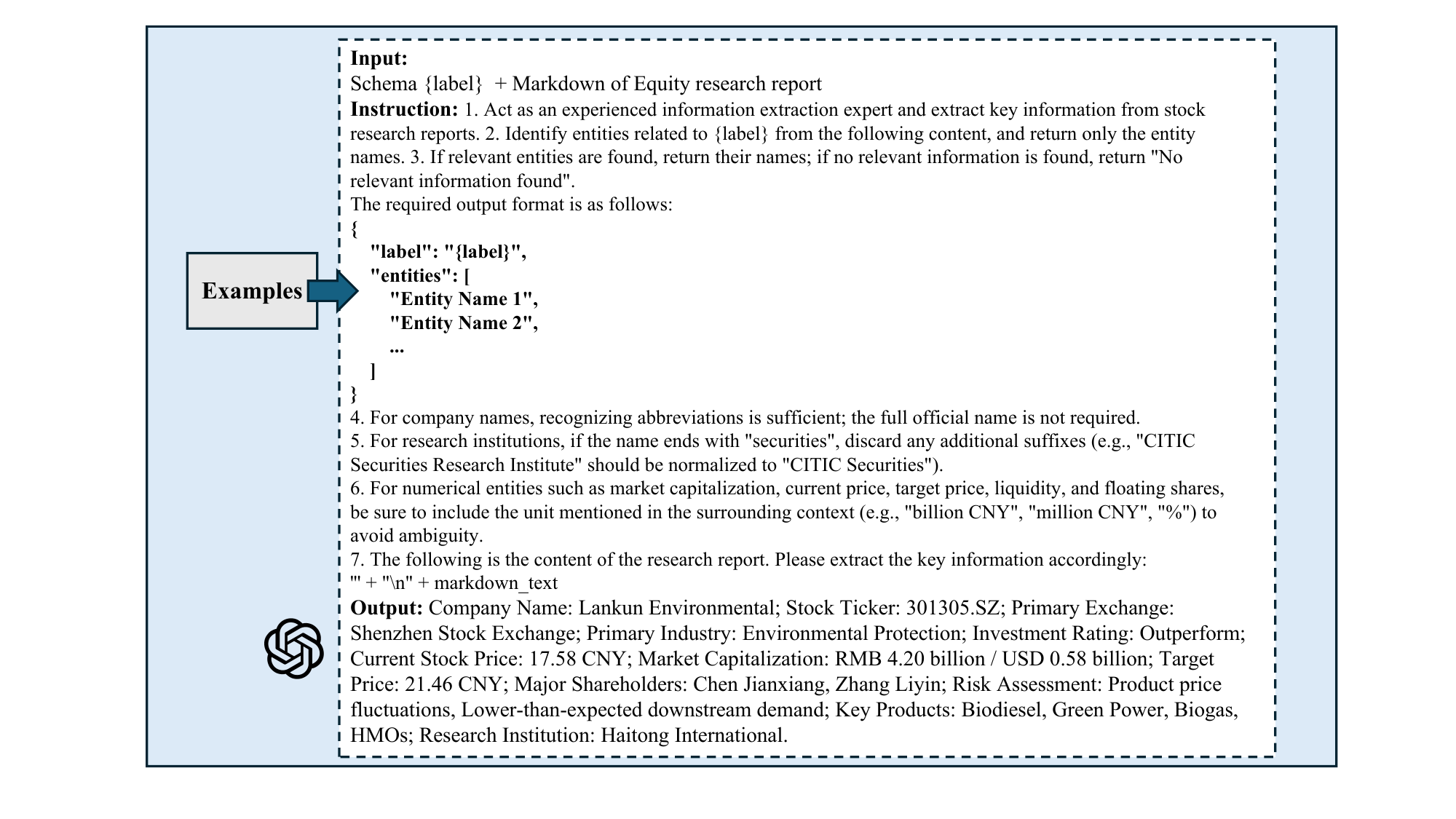}
  \caption{Prompt for Attribute graph construction.}
  \label{fig:attribute}
\end{figure*}

\subsection{The Prompt for Attribute Graph Construction}


After acquiring the schema of the Attribute Graph, this structure serves as the foundation for attribute-level knowledge population from financial research reports. Figure~\ref{fig:attribute} provides an example of the designed prompt used to guide the model in extracting these attributes during the knowledge population process.


\subsection{The Prompt for Acquiring Schema of Event Graph}

\begin{figure*}[ht]
  \centering
  \includegraphics[
    width=\textwidth,
    trim=10pt 2pt 5pt 10pt, 
    clip
  ]{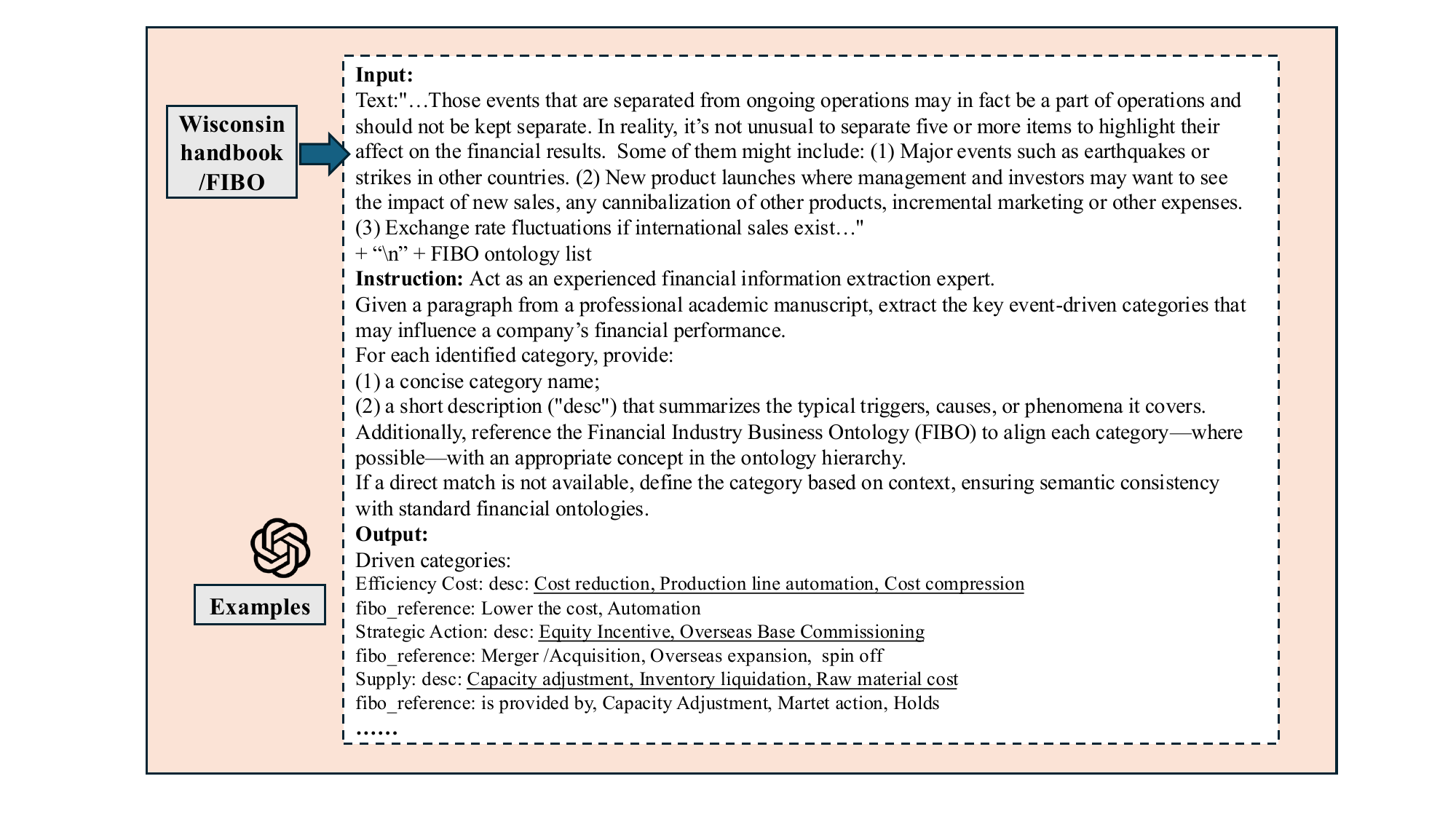}
  \caption{Prompt for acquiring schema of Event graph.}
  \label{fig:schema_event}
\end{figure*}

We automatically construct the top-down event-driven schema by prompting LLMs to extract high-level driven categories from the Wisconsin handbook and further match low-level event ontologies from FIBO to support interpretable event graph construction. Figure~\ref{fig:schema_event} illustrates the resulting tree-structured schema.
Each event category corresponds to a high-level driver and is associated with its fine-grained instances or relations. Below is the list of categories and their typical subtypes:

\begin{itemize}
    \item \textbf{Supply}: is provided by, Capacity Adjustment, Market Action, Holds
    \item \textbf{Demand}: Sales, Consumption, Performance, Is needed by
    \item \textbf{Revenue}: Earning, Profit, Income-oriented classifier, Is issued by, Has increased / decreased
    \item \textbf{Efficiency Cost}: Lower the cost, Automation
    \item \textbf{Strategic Action}: Merger / Acquisition, Overseas expansion, Spin-off
    \item \textbf{Technology Innovation}: Is applicable in, New product, Has innovated, Iteration
    \item \textbf{Policy Regulation}: Regulatory action, Governs, License
    \item \textbf{Macro}: Interest rate, GDP, Disaster
\end{itemize}

\subsection{The Prompt for Event Graph Construction}

\begin{figure*}[ht]
  \centering
  \includegraphics[
    width=\textwidth,
    trim=10pt 2pt 5pt 10pt, 
    clip
  ]{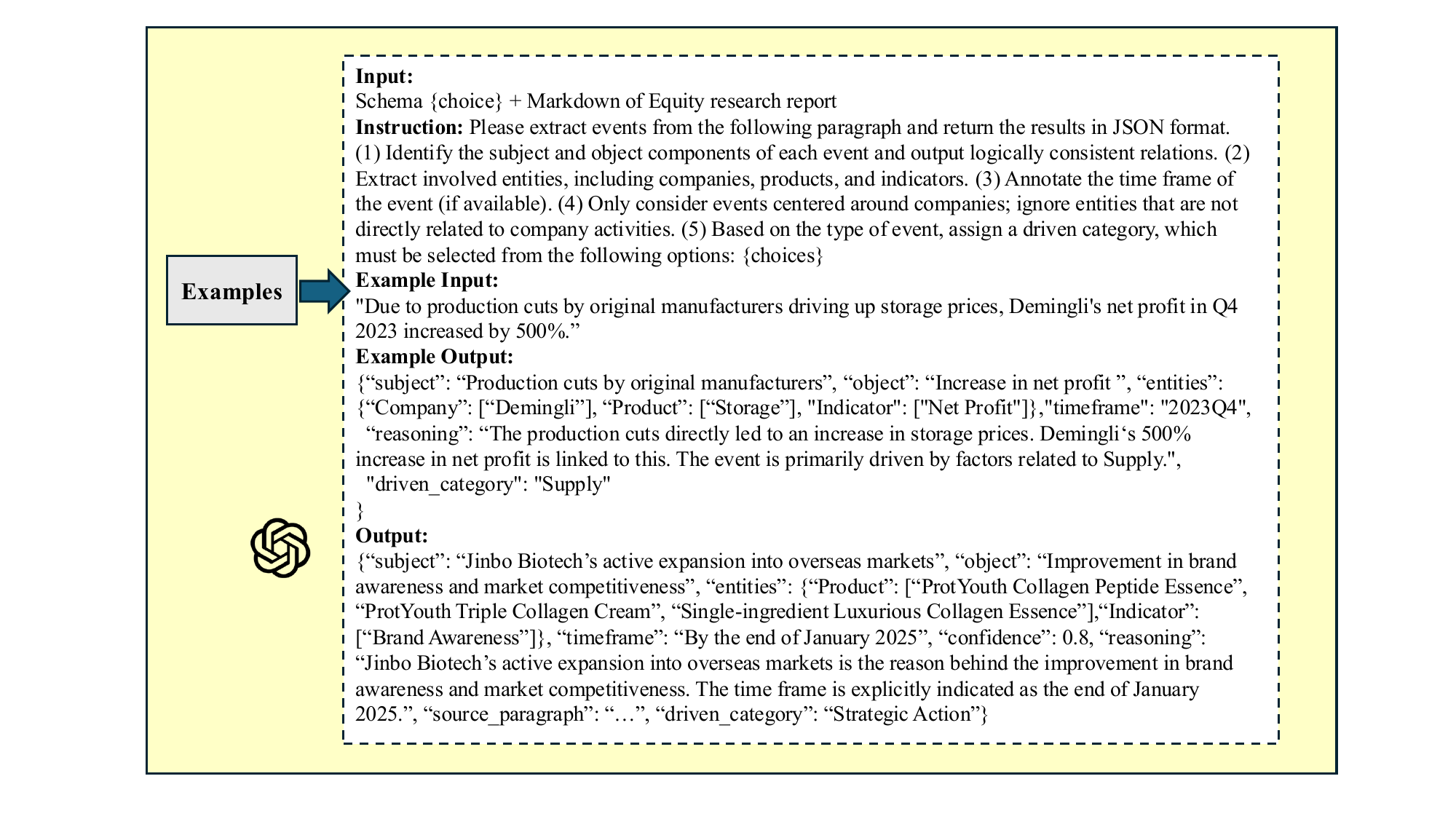}
  \caption{Prompt for Event graph construction.}
  \label{fig:event}
\end{figure*}

After establishing the event schema, we designed a tailored prompt to guide large language models in extracting structured event-level information from equity research reports. As illustrated in Figure~\ref{fig:event}, the prompt instructs the model to (1) identify the subject and object of the event, (2) extract relevant entities such as company names, products, and indicators, (3) annotate the timeframe, and (4) determine a driven category from options like ``Supply'', ``Demand'', or ``Strategic Action''.

To ensure relevance and accuracy, the prompt restricts extraction to events directly tied to company activities. The resulting JSON output includes not only the core event tuple but also a reasoning statement that explains the event linkage. The figure presents both an illustrative example and a real-case output to demonstrate the clarity and consistency achieved through our prompt design.



\section{The Case Study of Investment Query Response}

This subsection presents a case study comparing how various models—including FinLLMs such as FinGPT, XuanYuan-6B, Stock-Chain, and FinKario-RAG, as well as an advanced vanilla LLM (GPT-4o-mini), respond to a user query about investment analysis for Haier Biomedical (Haier Bio). The query asks the model to analyze the stock's investment potential and predict its future price trend.

\begin{figure*}[ht]
  \centering
  \includegraphics[
    width=\textwidth,
    trim=30pt 5pt 50pt 5pt, 
    clip
  ]{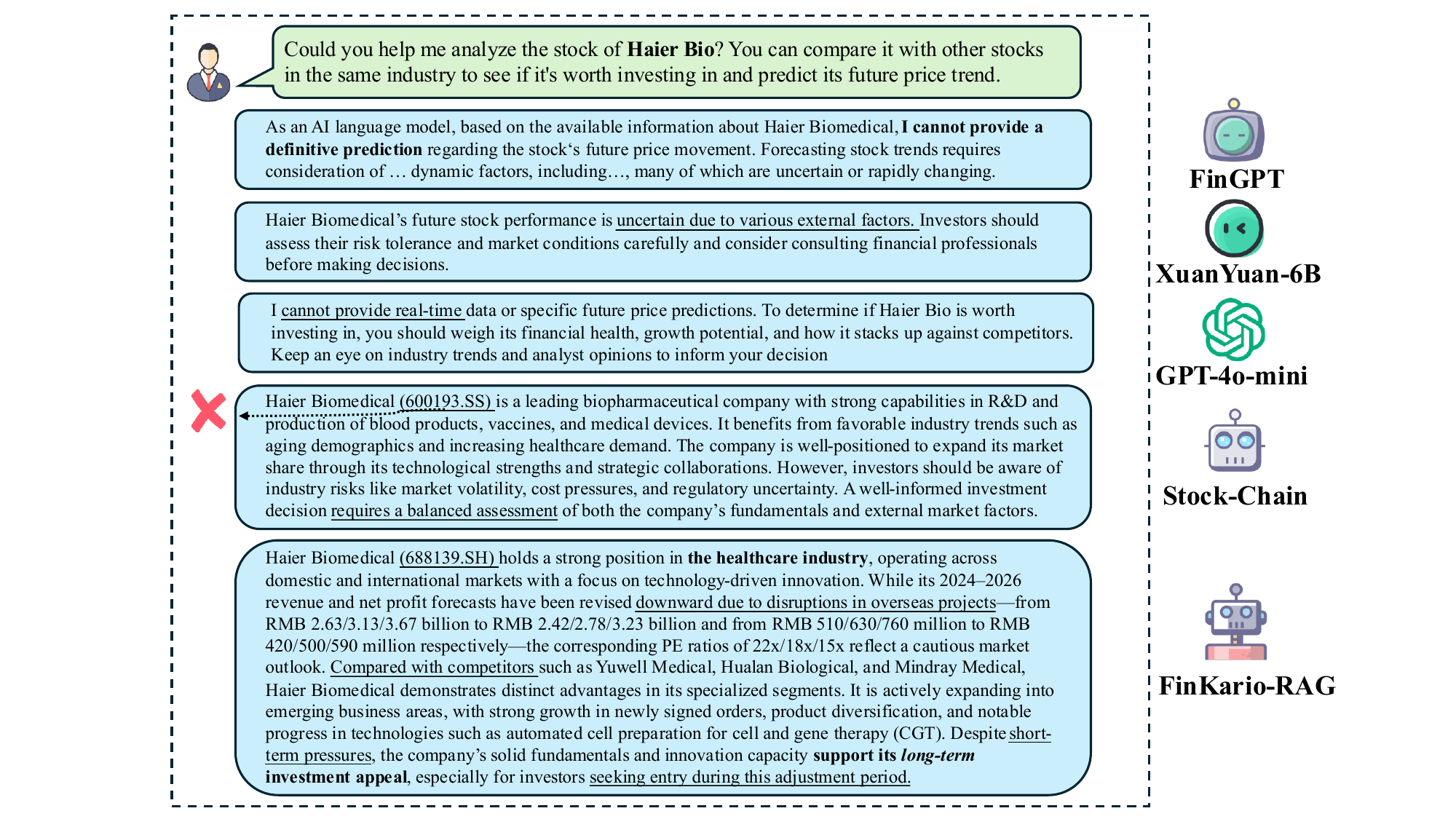}
  \caption{Evaluation of investment suggestions from FinGPT, XuanYuan-6B, GPT-4o-mini, Stock-Chain, and FinKario}
  \label{fig:qa}
\end{figure*}

As shown in Figure~\ref{fig:qa}, FinGPT, XuanYuan-6B, and GPT-4o-mini consistently emphasize the limitations of LLMs in delivering definitive investment predictions. These models cite the inherent uncertainty of market dynamics and the lack of access to real-time data as major barriers. Their responses generally avoid direct suggestions, instead encouraging users to consider macroeconomic conditions, company fundamentals, and to consult financial professionals.
In contrast, Stock-Chain attempts a more analytical response by summarizing company fundamentals and macro-level trends. However, its output includes factual inaccuracies such as misidentifying the stock code, and it fails to synthesize comparative insights across the industry, falling short of the user's request for a cross-company investment evaluation. Moreover, it does not offer actionable investment guidance, only suggesting that decisions require consideration of multiple external factors.

FinKario-RAG addresses these shortcomings by accurately grounding its analysis in correct entity identifiers, comparing the target stock with other industry players, and offering nuanced conclusions. This demonstrates the effectiveness of our financial knowledge graph construction and retrieval approach, which goes beyond traditional single-document retrieval methods to support more context-aware and investor-aligned responses.

\begin{figure*}[ht] 
    \centering
    \includegraphics[width=0.9\textwidth,
    trim=0pt 10pt 0pt 0pt, 
    clip]{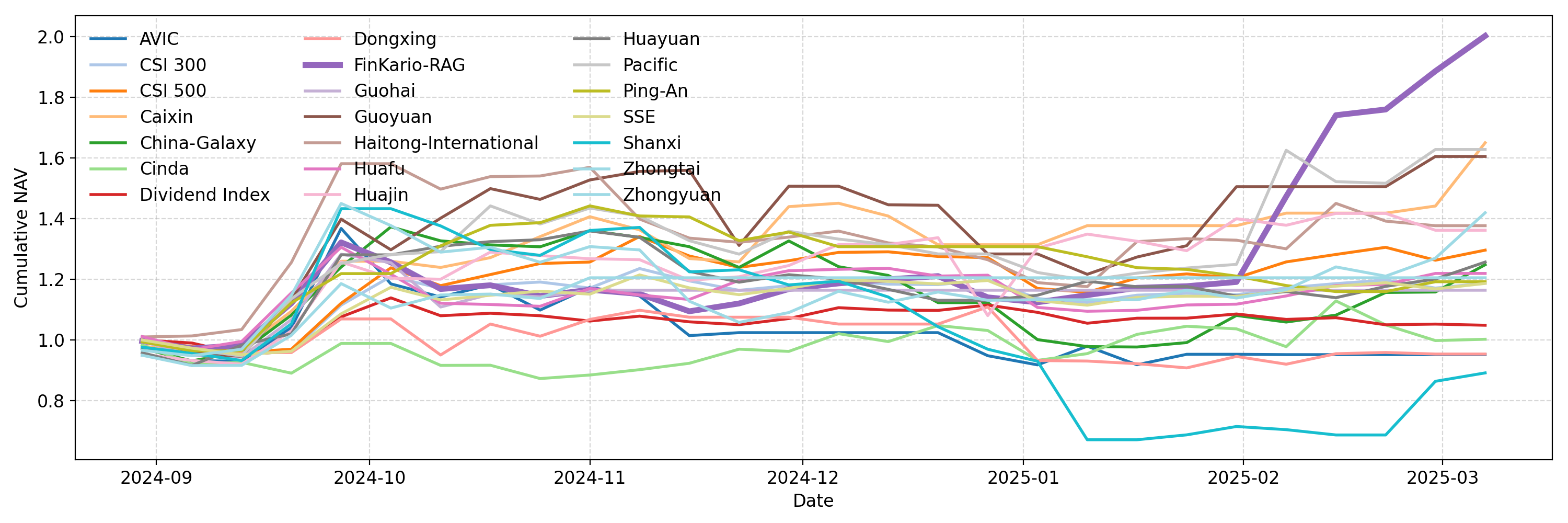} 
    \caption{{Accumulated returns (AR) of each institutional strategy on the financial‐report dataset from August 28, 2024 to March 7, 2025. The figure shows the net asset value (NAV) curves over the weekly backtesting period.}}
    \label{fig:nav_curve_100} 
    \vspace{-10px}
    
\end{figure*}

\section{The Supplementary Experiments of Institutional Strategy}

In the main manuscript, we compared our method against institutional agencies that published at least 300 equity research reports. To enable a more comprehensive evaluation of institutional strategy effectiveness, we additionally included agencies that have rated at least 100 reports in this supplementary analysis. The cumulative performance of these strategies is illustrated in the NAV curve shown in Figure~\ref{fig:nav_curve_100}, while detailed performance metrics are summarized in Table~\ref{tab:agency-results}.

For example, Guolian-Minsheng, SOOCHOW, and Caixin demonstrate strong annualized returns, with Guolian-Minsheng reaching 2.012, SOOCHOW at 1.625, and Caixin also at 1.625. Compared to low-return agencies such as Tianfeng and Cinda, whose annualized returns are below 0.06, these top performers yield more than 25 times higher returns. However, high returns do not always equate to stability. Caixin, despite having the highest Sharpe Ratio of 3.816, shows relatively low accuracy at 0.455, indicating that effective market timing can sometimes matter more than raw predictive accuracy. In contrast, Guolian-Minsheng achieves both a high return and an accuracy of 0.575, the best among all institutions. Zhongtai and Ping-An exhibit the lowest volatility values at 0.312 and 0.324, respectively, which are about 40\% lower than the average across institutions. Nonetheless, their returns remain moderate, indicating a trade-off between risk and reward. Meanwhile, Shanxi and Dongxing record negative returns, highlighting potential weaknesses in their investment strategies during weekly backtesting.
Some institutions, such as China-Post and Pacific, deliver reasonably high returns above 0.8 and 1.5, respectively. However, their performance metrics such as drawdown and accuracy remain suboptimal, indicating inconsistent prediction quality. Agencies like Guosen and KaiYuan exhibit more balanced profiles with moderate returns and volatility, yet lack standout performance in any single dimension.

Overall, these results illustrate the fragmented quality and strategic effectiveness among institutional players. Against this backdrop, FinKario-RAG delivers the strongest performance across all major return-oriented metrics, including the highest annualized return, Sharpe Ratio, and accuracy, while keeping risk measures such as maximum drawdown and volatility within acceptable bounds, thereby reinforcing the advantage of our structured retrieval and knowledge-grounded approach.

\begin{table*}[htbp]
\centering
\caption{Performance of institutional agencies that published at least 100 equity research reports.}
\label{tab:agency-results}
\renewcommand{\arraystretch}{0.8}
\setlength{\tabcolsep}{16pt}
\begin{tabular}{lrrrrrr}
\toprule
\textbf{Agency} & \textbf{ARR↑} & \textbf{VOL↓} & \textbf{SR↑} & \textbf{MDD↓} & \textbf{CR↑} & \textbf{ACC↑} \\
\midrule
Tianfeng            & 0.054 & 0.542 & 0.100 & 0.225 & 0.242 & 0.411 \\
Southwest           & 0.121 & 0.485 & 0.249 & 0.173 & 0.701 & 0.492 \\
Sinolink            & 0.391 & 0.648 & 0.604 & 0.365 & 1.070 & 0.438 \\
SOOCHOW             & 1.625 & 0.522 & \cellcolor[rgb]{ .988, .89, .902}3.115 & 0.132 & \cellcolor[rgb]{ .988, .89, .902}12.311 & 0.557 \\
Guolian-Minsheng    & \cellcolor[rgb]{ .98, .635, .643}2.012 & 0.647 & 3.108 & 0.169 & 11.880 & \cellcolor[rgb]{ .98, .635, .643}0.575 \\
Guosen              & 0.167 & 0.456 & 0.366 & 0.197 & 0.845 & 0.460 \\
Huaan               & 0.170 &0.471 & 0.361 & 0.333 & 0.509 & 0.435 \\
KaiYuan             & 0.181 & 0.473 & 0.383 & 0.279 & 0.650 & 0.552 \\
China-Fortune       & 0.263 & 0.537 & 0.489 & 0.216 & 1.218 & \cellcolor[rgb]{ .988, .89, .902}0.573 \\
China-Post          & 0.830 & 0.559 & 1.485 & 0.236 & 3.519 & 0.440 \\
Pacific          & \cellcolor[rgb]{ .988, .89, .902}1.557 & 0.582 & 2.674 & 0.170 & 9.138 & 0.458 \\
China-Galaxy          & 0.533 & 0.452 & 1.179 & 0.289 & 1.847 & 0.482\\
Huafu          & 0.465 & \cellcolor[rgb]{ .988, .89, .902}0.381 & 1.222 & 0.162 & 2.881 & 0.409 \\
Zhongtai          & 0.432 & \cellcolor[rgb]{ .973, .412, .42}0.312 & 1.388 & \cellcolor[rgb]{ .973, .412, .42}0.068 & 6.358 & 0.328 \\
Cinda          & 0.005 & 0.413 & 0.011 & \cellcolor[rgb]{ .98, .635, .643}0.117 & 0.041 & 0.496 \\
Shanxi          & -0.198 & 0.817 & -0.243 & 0.532 & -0.373 & 0.440 \\
Guohai          & 0.340 & 0.387 & 0.879 & 0.136 & 2.506 & 0.402 \\
Huajin          & 0.813 & 0.569 & 1.429 & 0.192 & 4.233 & 0.358 \\
Ping-An          & 0.403 & \cellcolor[rgb]{ .98, .635, .643}0.324 & 1.243 & 0.196 & 2.056 & 0.538 \\
Guoyuan          & 1.488 & 0.623 & 2.391 & 0.220 & 6.768 & 0.491 \\
Huayuan         & 0.553 & 0.421 & 1.314 & 0.168 & 3.292 & 0.409 \\
Haitong-International          & 0.852 & 0.571 & 1.490 & 0.256 & 3.322 & 0.395 \\
Zhongyuan          & 0.964 & 0.561 & 1.718 & 0.270 & 3.565 & 0.483 \\
Caixin          & 1.625 & 0.426 & \cellcolor[rgb]{ .98, .635, .643}3.816 & \cellcolor[rgb]{ .988, .89, .902}0.126 & \cellcolor[rgb]{ .98, .635, .643}12.897 & 0.455 \\
AVIC          & -0.091 & 0.589 & -0.154 & 0.329 & -0.276 & 0.437 \\
Dongxing          & -0.087 & 0.391 & -0.223 & 0.181 & -0.481 & 0.415 \\

\midrule\midrule
\textbf{FinKario-RAG} & \cellcolor[rgb]{ .973, .412, .42}2.633 & 0.534 & \cellcolor[rgb]{ .973, .412, .42}4.926 & 0.172 & \cellcolor[rgb]{ .973, .412, .42}15.315 & \cellcolor[rgb]{ .973, .412, .42}0.581 \\
\bottomrule
\end{tabular}
\end{table*}

\end{document}